\documentclass[journal]{IEEEtran}
\usepackage{amsmath,amsfonts, amssymb}
\usepackage{algorithmic}
\usepackage{array}
\usepackage[caption=false,font=normalsize,labelfont=sf,textfont=sf]{subfig}
\usepackage{textcomp}
\usepackage{stfloats}
\usepackage{url}
\usepackage{verbatim}
\usepackage{graphicx}
\usepackage{adjustbox}  

\usepackage{booktabs, multirow, makecell}
\usepackage[ruled]{algorithm2e}
\usepackage{amsthm}
\usepackage[colorlinks, linkcolor=blue, anchorcolor=blue, citecolor=blue]{hyperref}
\usepackage{cleveref}
\crefname{equation}{Eq.}{Eqs.}
\Crefname{equation}{Equation}{Equations}

\usepackage{orcidlink}

\newtheorem{theorem}{Theorem}
\newtheorem{property}{Property}
\newtheorem{definition}{Definition}

\usepackage{balance}

\begin{document}
\title{TV Subgradient-Guided Multi-Source Fusion for Spectral Imaging \\in Dual-Camera CASSI Systems}
\author{Weiqiang Zhao\orcidlink{0009-0009-2714-4622},
Tianzhu Liu\orcidlink{0000-0001-6903-9614}, \IEEEmembership{Member, IEEE},
Yuzhe Gui\orcidlink{0009-0007-2630-6736},
Wei Bian\orcidlink{0000-0003-4252-047X},
Yanfeng Gu\orcidlink{0000-0003-1625-7989}, \IEEEmembership{Senior Member, IEEE}
\thanks{This work was supported by the National Science Fund for Distinguished Young Scholars 62025107 (Corresponding author: Yanfeng Gu).}
\thanks{Weiqiang Zhao, Tianzhu Liu, Yuzhe Gui, Yanfeng Gu are with the School of Electronics and Information Engineering, Harbin Institute of Technology, Harbin 150001, China (E-mail: wqzhao@stu.hit.edu.cn; tzliu@hit.edu.cn; 25B305002@stu.hit.edu.cn; guyf@hit.edu.cn).}
\thanks{Wei Bian is with the School of Mathematics, Harbin Institute of Technology, Harbin 150001, China (E-mail: bianweilvse520@163.com).
}}

\markboth{Journal of \LaTeX\ Class Files,~Vol.~18, No.~9, September~2020}
{Weiqiang Zhao, Tianzhu Liu, \MakeLowercase{\textit{(et al.)}}: TV Subgradient-Guided Multi-Source Fusion for Spectral Imaging in Dual-Camera CASSI Systems}

\maketitle
\IEEEpubid{\begin{minipage}{\textwidth}\ \\[30pt] \centering 
Copyright \copyright 2026 IEEE. Personal use of this material is permitted. \\
However, permission to use this material for any other purposes must be obtained from the IEEE by sending an email to pubs-permissions@ieee.org.
\end{minipage}}

\begin{abstract}
    Balancing spectral, spatial, and temporal resolutions is a key challenge in spectral imaging. The Dual-Camera Coded Aperture Snapshot Spectral Imaging (DC-CASSI) system alleviates this trade-off but suffers from severely ill-posed reconstruction problems due to its high compression ratio. Existing methods are constrained by scene-specific tuning or excessive reliance on paired training data. To address these issues, we propose a Total Variation (TV) subgradient-guided multi-source fusion framework for DC-CASSI reconstruction, comprising three core components: (1) An end-to-end Single-Disperser CASSI (SD-CASSI) observation model based on the tensor-form Kronecker $\delta$, which establishes a rigorous mathematical foundation for physical constraints while enabling efficient adjoint operator implementation; (2) An adaptive spatial reference generator that integrates SD-CASSI’s physical model and RGB subspace constraint, generating the reference image as reliable spatial prior; (3) A TV subgradient-guided regularization term that encodes local structural directions from the reference image into spectral reconstruction, achieving high-quality fused results. The framework is validated on simulated datasets and real-world datasets. Experimental results demonstrate that it achieves state-of-the-art reconstruction performance and robust noise resilience. This work not only establishes an interpretable theoretical foundation for subgradient-guided fusion but also provides a practical fusion-based paradigm for high-fidelity spectral image reconstruction in DC-CASSI systems. Source code: https://github.com/bestwishes43/ADMM-TVDS.
\end{abstract}

\begin{IEEEkeywords}
Snapshot compressive imaging, image fusion, subgradient, total variation
\end{IEEEkeywords}

\section{Introduction}
\IEEEPARstart{S}{pectral} imaging has become an indispensable tool for quantitative characterization of objects, enabling precise analysis of chemical compositions, physical properties, and biological traits by capturing spatial-spectral data cubes. 
Its applications span critical fields such as remote sensing, geological exploration, and medical diagnosis. However, as {highlighted in} \cite{bian_broadband_2024}, conventional single-sensor scanning-based spectral systems face a fundamental trade-off among spatial, spectral, and temporal resolutions.

To address this limitation, computational spectral imaging (CSI) has emerged as a transformative paradigm. It leverages compressed sensing for single-shot spatial–spectral acquisition via optical encoding and algorithmic reconstruction \cite{SpectralImagingWithDeepLearning}. Among CSI methods, coded aperture snapshot spectral imaging (CASSI) dominates, compressing a 3D spectral cube into a 2D measurement using a coded aperture and a dispersive element \cite{SD_CASSI,DD-CASSI}. Recent SD-CASSI reconstruction advancements \cite{feng_2d-slice_2025, li_multi-feature_2025} have adopted deep-unfolding frameworks embedding physical observation priors, but rely on the computationally inefficient two-stage observation model. More critically, the aggressive compression in SD-CASSI renders the inverse problem severely ill-posed, leaving insufficient constraints for stable single-camera recovery.

This has spurred dual-camera CASSI (DC-CASSI), which augments CASSI with an auxiliary RGB or panchromatic camera \cite{DC-CASSI_pan,DC-CASSI-RGB}. Early DC-CASSI relied on data-level fusion, neglecting cross-modal complementarity \cite{DC-CASSI_pan,DC-CASSI-RGB, rTVRA}; recent work adopts feature-level fusion. Yet current approaches remain limited: supervised methods require sensor-matched paired data and generalize poorly \cite{In2SET_2024, CasFormer,MLP-AMDC_2025}; self-supervised ones trade robustness for high computational cost \cite{PiE,SIGDU-Net}; and model-based techniques demand challenging hardware configuration \cite{PFusion} or scene-specific tuning \cite{ADMM_PIDS}.

\begin{figure*}[ht]
    \centering
    \includegraphics[width=\textwidth]{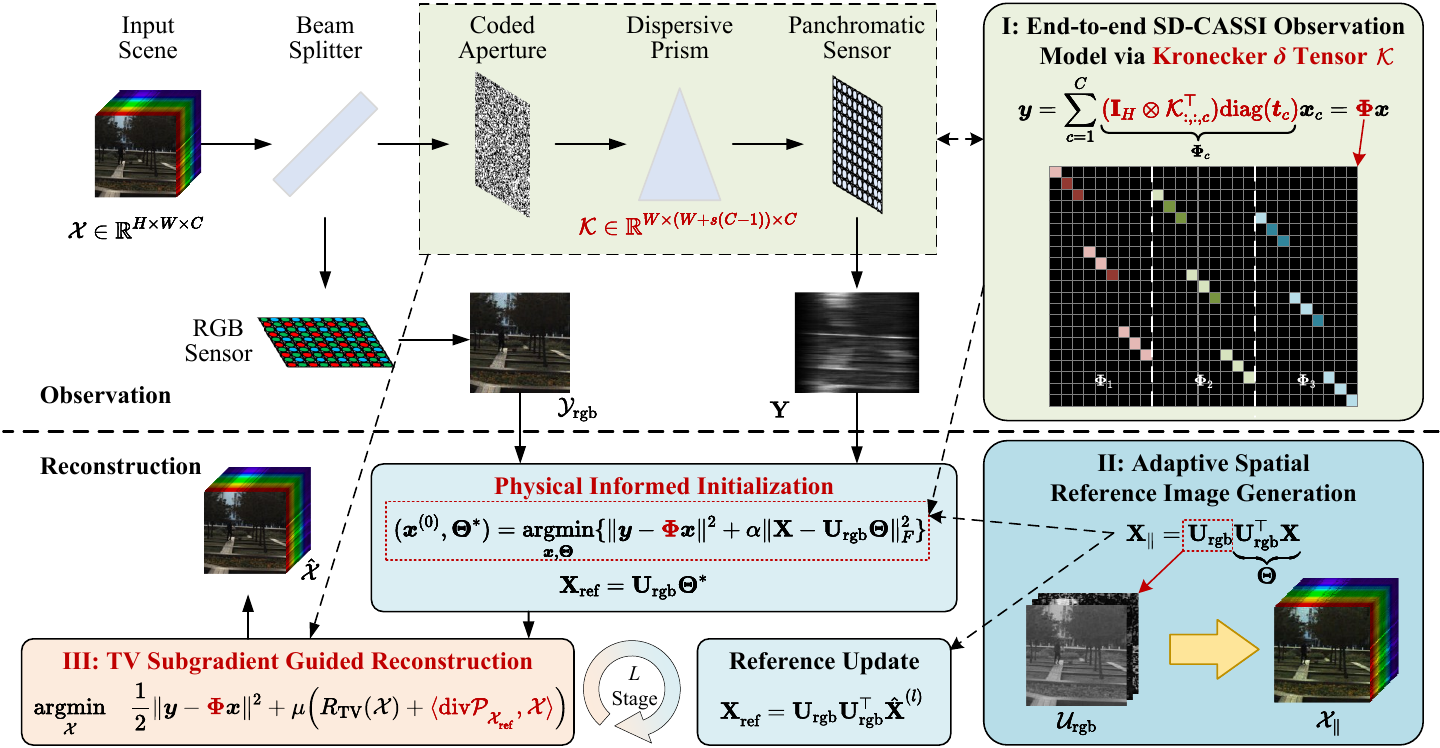}
    \caption{Proposed TV subgradient-guided multi-source fusion framework for DC-CASSI}
    \label{Figure:SystemModule}
\end{figure*}

To tackle these challenges, we propose a model-based, multi-source fusion framework for DC-CASSI, as illustrated in \Cref{Figure:SystemModule}. The framework consists of three tightly integrated modules, each contributing a distinct innovation:

\begin{itemize}
    \item \textbf{End-to-end SD-CASSI observation modeling}: We formulate a tensor-form Kronecker $\delta$ based observation model, which rigorously encodes the spatial-spectral coding physics and enables an efficient adjoint operator. 
    \item \textbf{Adaptive spatial reference generation}: We develop a reference generation strategy that jointly exploits the SD-CASSI observation model and the RGB subspace structure to extract and refine spatial priors. It eliminates the need for scene- or sensor-specific parameter tuning.
    \item \textbf{TV subgradient-guided regularization}: We propose a regularization term (named TVDS) based on subgradient theory. This term aligns the spatial feature characterized by the TV subgradient of the reconstructed image with that of the RGB-derived reference image, thereby fully leveraging the complementary information of DC-CASSI.
\end{itemize}

Experiments on both simulated and real datasets confirm that the proposed method performs well in reconstruction quality, noise robustness, and computational efficiency.

The remainder of this paper is organized as follows: \Cref{sec2} reviews related works on CASSI modeling and reconstruction. \Cref{sec:System_Model} and \Cref{Init} elaborate on the DC-CASSI observation model and adaptive spatial reference generation, respectively. \Cref{sec:TVDS} presents the TV subgradient-guided reconstruction algorithm and its ADMM implementation. \Cref{sec6} validates the method on simulated datasets, while \Cref{sec7} extends the evaluation to real datasets. \Cref{sec8} concludes the study.

\section{Related Works}
\label{sec2}
\noindent For DC-CASSI reconstruction, the MAP estimation framework admits two formulations:
\begin{align}
    \underset{\mathcal{X}}{\mathrm{argmax}} \quad & \log P(\mathbf{Y}|\mathcal{X}) + \log P(\mathbf{Y}_{\text{aux}}|\mathcal{X}) + \log P(\mathcal{X}), \label{P1}\\
    \underset{\mathcal{X}}{\mathrm{argmax}} \quad & \log P(\mathbf{Y}|\mathcal{X}) + \log P(\mathcal{X}|\mathbf{Y}_{\text{aux}}), \label{fusionBayesian}
\end{align} 
where $\mathbf{Y}$ and $\mathbf{Y}_{\text{aux}}$ represent observations from the CASSI system and auxiliary camera, respectively, and $\mathcal{X}$ denotes the spectral image to be reconstructed. 
Specifically, \cref{P1} combines dual data fidelity constraints with intrinsic image priors, while \cref{fusionBayesian} achieves feature-level fusion via conditional priors from auxiliary data. Solving these estimation problems requires modeling the CASSI observation process and designing reconstruction priors, both reviewed subsequently.

\subsection{Modeling of CASSI Observation Process}

CASSI systems compress spectral images into 2D measurements via spatial-spectral joint coding. While early dual-disperser configuration \cite{DD-CASSI} achieve 3D coding, their stringent optical alignment requirements hindered practical applications. To overcome this limitation, single-disperser CASSI (SD-CASSI) \cite{SD_CASSI} emerged as the mainstream configuration by simplifying the optical path through removing the first disperser.

The SD-CASSI observation process follows a linear model $\boldsymbol{y}=\mathbf{\Phi}\boldsymbol{x}$, where $\mathbf{\Phi}$ is the observation matrix, $\boldsymbol{y}$ is the vectorized measurement, and $\boldsymbol{x}$ is the vectorized spectral image. Early studies \cite{SD_CASSI, SD_CASSI_DAmici} characterized this process in an element-wise manner, whereas subsequent works \cite{josaa_2011, tip_2013} proposed overly complex observation matrices that are impractical for reconstruction algorithms.

To address this challenge, a two-stage modeling scheme has been proposed. This approach first disperse original spectral image and then establishes the linear relationship $\boldsymbol{y}=\mathbf{\Phi}'\boldsymbol{x}'$, where $\mathbf{\Phi}'$ is the mapping matrix, and $\boldsymbol{x}'$ is the vectorized dispersed spectral image. 
The simple structure of $\mathbf{\Phi}'$ \cite{PNP-CASSI} enables efficient derivation and implementation of the adjoint operator of $\mathbf{\Phi}'$, which is essential in reconstruction. Consequently, this two-stage scheme has been widely adopted in both model-based \cite{NLRT} and deep unfolding methods \cite{GAP-Net, DAUHST, feng_2d-slice_2025, li_multi-feature_2025}.

To establish a direct end-to-end mapping, researchers \cite{spm_2014, HyperReconNet, PNP-SCI} have attempted to remove zero-padded regions (introduced via dispersion) from $\boldsymbol{x}'$ and corresponding columns in $\mathbf{\Phi}'$. However, this model lacks a rigorous derivation, hindering the analysis of adjoint operator for efficient implementation. As a result, researchers \cite{TSA_Net, RDLUF, In2SET_2024, ADMM_PIDS, PiE, SIGDU-Net} often use the end-to-end model for problem formulation but implement methods based on the two-stage framework.

To date, no mathematically rigorous end-to-end observation model of SD-CASSI has been proposed, which restricts theoretical analyses of reconstruction techniques.

\subsection{Reconstruction Methods of CASSI}
The reconstruction methods are {designed to address} the ill-posed inverse problem through priors constraints, which can be categorized into three technical paradigms:

Learning-based methods employ neural networks to capture spectral image statistics and structure (e.g., HyperReconNet \cite{HyperReconNet}, TSA-Net \cite{TSA_Net}, MST \cite{MST}). In DC-CASSI, multi-source fusion approaches dominate \cite{ijcv_2024, MLP-AMDC_2025, CasFormer}, but supervised variants suffer from dependency on paired training data (difficult to obtain) and poor generalization on sensors and scenes. Self-supervised methods (e.g., PiE \cite{PiE}, SIGDU-Net \cite{SIGDU-Net}) address these problems but {they are} time-consuming.

Deep unfolding networks integrate physics-driven optimization with learned proximal operators (e.g., \cite{GAP-Net, ADMM_CSNet, DAUHST, In2SET_2024}), embedding system priors explicitly. These methods require fewer training samples, support adjustable system configurations, and maintain interpretability. However, their "black-box" deep learning components lack rigorous theoretical guarantees for convergence and stability.

Model-based methods {employ} intrinsic priors such as sparsity \cite{SD_CASSI, SD_CASSI_sparse}, low-rankness \cite{DeSCI, DDLR_2019}, and smoothness \cite{GAP_TV, kittle_multiframe_2010} for spectral reconstruction. However, when extended to DC-CASSI \cite{DC-CASSI_pan, DC_NLSR, rTVRA} as in \cref{P1}, auxiliary observations provide limited information because they constrain only data fidelity. The \cref{fusionBayesian} paradigm improves performance via feature-fusion but suffers from individual limitations: PFusion \cite{PFusion} fuses dual-disperser CASSI and RGB in low-rank subspace but requires impractical dual-disperser configuration; PIDS \cite{ADMM_PIDS} fuses spatial difference feature from prior image into SD-CASSI reconstruction but is sensitive to energy alignment parameter between prior image and ground truth; ADMM-PRM \cite{ADMM-PRM} addresses PIDS limitations with random forest-based prior generation but severely compromises computational efficiency.

\section{Observation Model of DC-CASSI}
\label{sec:System_Model}
{
\noindent As a typical multi-spectral imaging sensor system, the DC-CASSI architecture integrates compressive sensing based SD-CASSI and an RGB/panchromatic camera, plus a beam splitter to enable synchronous dual-source acquisition as illustrated in \Cref{Figure:SystemModule}. To establish a rigorous physical basis for subsequent multi-source fusion (e.g., data-level consistency constraints in \Cref{Init} and feature-level guided fusion in \Cref{sec:TVDS}), this section analyzes the observation models of the two sensors. Specifically, we focus on the SD-CASSI + RGB configuration, as it is the most widely used DC-CASSI system.}

\subsection{Observation Model of SD-CASSI System}
\begin{figure}[htbp]
    \centering
    \includegraphics[width=0.95\linewidth]{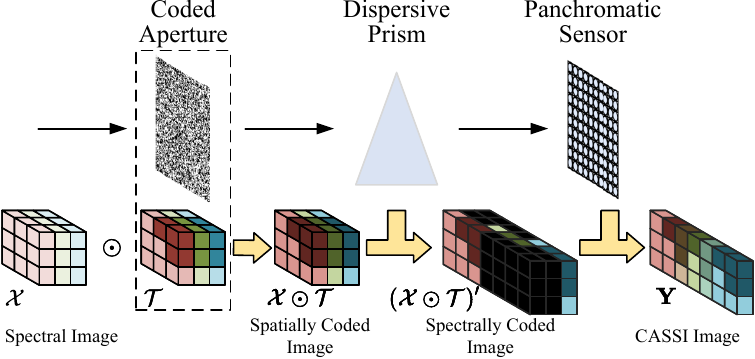}
    \caption{Visualization of observation process of SD-CASSI}
    \label{Visualization:Phi}
\end{figure}

{
As shown in \Cref{Visualization:Phi}, the incident light first undergoes spatial modulation via a coded aperture. The modulated light is subsequently dispersed by a prism with a shear step of $s\in\mathbb{N}^+$, yielding a spectrally sheared image. This dispersed encoded spectral information is finally integrated by the panchromatic sensor, forming the CASSI image. According to \cite{SpectralImagingWithDeepLearning}, the simplified discrete observation model can be formulated as
\begin{equation}
    \mathbf{Y} = \sum_{c=1}^{C} (\mathcal{X} \odot \mathcal{T})'_{:,:,c},
    \label{Equation:Former_Observation}
\end{equation}
where $\mathcal{X}\in\mathbb{R}^{H\times W\times C}$ denotes the spectral image of the target scene, $\mathbf{Y}\in\mathbb{R}^{H\times (W+s(C-1))}$ is the SD-CASSI image, and $\mathcal{T}\in\mathbb{R}^{H\times W\times C}$ represents the equivalent coded aperture mask that incorporates the influences of the split ratio, original coded aperture, and quantum efficiency of the panchromatic sensor. $\odot$ denotes the Hadamard product, and $(\mathcal{X}\odot\mathcal{T})'\in \mathbb{R}^{H\times(W+s(C-1))\times C}$ represents the shear-transformed version of $\mathcal{X}\odot\mathcal{T}$, which is defined as
\begin{equation*}
    (\mathcal{X} \odot \mathcal{T})'_{:,:,c} = \begin{bmatrix}
        \mathbf{0}_{H\times s(c-1)} & (\mathcal{X} \odot \mathcal{T})_{:,:,c} & \mathbf{0}_{H\times s(C-c)}
    \end{bmatrix} ,
\end{equation*}}

Obviously, the shear transform can be swapped with the Hadamard product. In prior studies \cite{TSA_Net, NLRT, GAP-Net, ADMM-PRM}, researchers have introduced the shear-transformed spectral image $\mathcal{X}'\in\mathbb{R}^{H\times (W+s(C-1))\times C}$ and shear-transformed mask $\mathcal{T}'\in\mathbb{R}^{H\times (W+s(C-1))\times C}$ to model the SD-CASSI observation process, which are defined as
\begin{equation*}
    \begin{aligned}
    \mathcal{X}'_{:,:,c} &= \begin{bmatrix}
        \mathbf{0}_{H\times s(c-1)} & \mathcal{X}_{:,:,c} & \mathbf{0}_{H\times s(C-c)} 
    \end{bmatrix}, \\
    \mathcal{T}'_{:,:,c} &= \begin{bmatrix}
        \mathbf{0}_{H\times s(c-1)} & \mathcal{T}_{:,:,c} & \mathbf{0}_{H\times s(C-c)} 
    \end{bmatrix}, \\
    \end{aligned} \label{Equation:Shifted_Form}
\end{equation*}
this transforms \cref{Equation:Former_Observation} into
\begin{equation*}
    \mathbf{Y} = \sum_{c=1}^{C} \mathcal{X}'_{:,:,c} \odot \mathcal{T}'_{:,:,c},
    \label{Equation:Former_Observation_Shifted}
\end{equation*}
on this basis, the observation model is constructed in a vectorized form as follows:
\begin{equation}
    \boldsymbol{y} = \sum_{c=1}^{C} \mathrm{diag}(\boldsymbol{t}'_c)\boldsymbol{x}'_c = \mathbf{\Phi}' \boldsymbol{x}',
    \label{Equation:Phi_ori}
\end{equation} 
where \(\mathbf{\Phi}'=\left[\text{diag}(\boldsymbol{t}'_1), \text{diag}(\boldsymbol{t}'_2), \cdots, \text{diag}(\boldsymbol{t}'_C)\right]\) is observation matrix, \(\boldsymbol{x}'=[\boldsymbol{x}'^\top_1, \boldsymbol{x}'^\top_2, \cdots, \boldsymbol{x}'^\top_C]^\top\) is the vectorized \(\mathcal{X}'\), and $\boldsymbol{y}$, $\boldsymbol{x}'_c$, and $\boldsymbol{t}'_c$ are the vectorized forms of $\mathbf{Y}$, $\mathcal{X}'_{:,:,c}$, and $\mathcal{T}'_{:,:,c}$, respectively. $\text{diag}(\cdot)$ denotes a square diagonal matrix with the elements of the input vector on the main diagonal. Additionally, some studies \cite{SIGDU-Net,ADMM_PIDS, PiE} remove the rows corresponding to the padded zeros in $\boldsymbol{x}'$ (denoted as $\boldsymbol{x}$) and the corresponding columns in $\mathbf{\Phi}'$ (denoted as $\mathbf{\Phi}$) to obtain the end-to-end observation model $\boldsymbol{y}=\mathbf{\Phi}\boldsymbol{x}$. However, this approach hinders efficient adjoint operator analysis because it breaks the corresponding relationship between the vector-form and tensor-form observation models.

\begin{figure}[htbp]
    
    \centering
    \adjustbox{padding=3pt}{
        \includegraphics[width=0.95\linewidth]{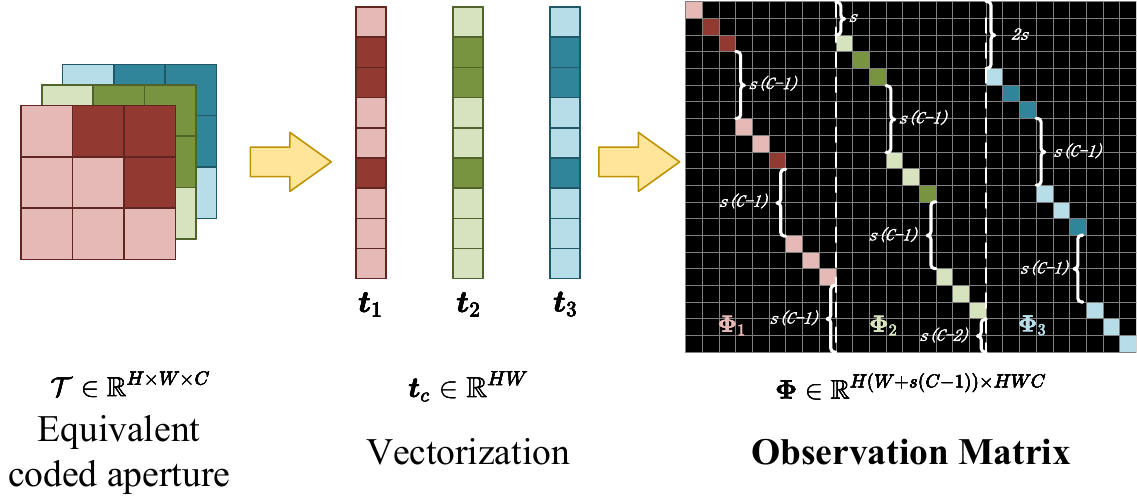}
    }
    \caption{Visualization of proposed SD-CASSI observation matrix}
    \label{Visualization:SD-CASSI-Observation}
\end{figure}

Different from aforementioned studies, we describe the shear transform using the tensor-form Kronecker delta $\mathcal{K}\in\mathbb{R}^{W\times (W + s(C-1))\times C}$, which is defined as
\begin{equation*}
   \mathcal{K}_{:,:,c} = \begin{bmatrix} \mathbf{0}_{W\times s(c-1)} & \mathbf{I}_{W\times W} & \mathbf{0}_{W\times s(C-c)} \end{bmatrix},
\end{equation*}
using this definition, \cref{Equation:Former_Observation} can be reformulated as:
\begin{equation}
    \begin{aligned}
        \mathbf{Y} = \sum_{c=1}^{C} (\mathcal{X}_{:,:,c}\odot\mathcal{T}_{:,:,c}){\mathcal{K}}_{:,:, c},
    \end{aligned}
    \label{Equation:Encoder_Sum}
\end{equation}
by vectorizing \cref{Equation:Encoder_Sum}, we obtain the end-to-end observation model of SD-CASSI as:
\begin{equation}
  \boldsymbol{y} = \sum_{c=1}^{C} \underbrace{(\mathbf{I}_H \otimes \mathcal{K}_{:, :, c}^\top) \mathrm{diag}(\boldsymbol{t}_{c})}_{\mathbf{\Phi}_c}\boldsymbol{x}_{c} = \mathbf{\Phi} \boldsymbol{x}.
  \label{Equation:CASSI_Vec}
\end{equation} 
where $\boldsymbol{x}_c$ and $\boldsymbol{t}_c$ are the vectorized forms of $\boldsymbol{\mathcal{X}}_{:, :, c}$ and $\boldsymbol{\mathcal{T}}_{:, :, c}$, respectively, and $\otimes$ denotes the Kronecker product. Let $\mathbf{\Phi}_c = (\mathbf{I}_H \otimes \mathcal{K}_{:, :, c}^\top) \mathrm{diag}(\boldsymbol{t}_c)$, the sensing matrix is $\mathbf{\Phi}=\left[\mathbf{\Phi}_1, \mathbf{\Phi}_2, \cdots ,\mathbf{\Phi}_C\right]$, and $\boldsymbol{x}=[\boldsymbol{x}^\top_1, \boldsymbol{x}^\top_2, \cdots, \boldsymbol{x}^\top_C]^\top$. {For a better understanding of the proposed observation model, we visualize the observation matrix in \Cref{Visualization:SD-CASSI-Observation}}.

\subsection{Discussions on Proposed SD-CASSI Observation Model}
For the proposed SD-CASSI observation model \(\boldsymbol{y} = \mathbf{\Phi}\boldsymbol{x}\), it possesses two properties that are convenient for reconstruction.

\begin{figure}[htbp]
    \centering
    
    \includegraphics[width=0.95\linewidth]{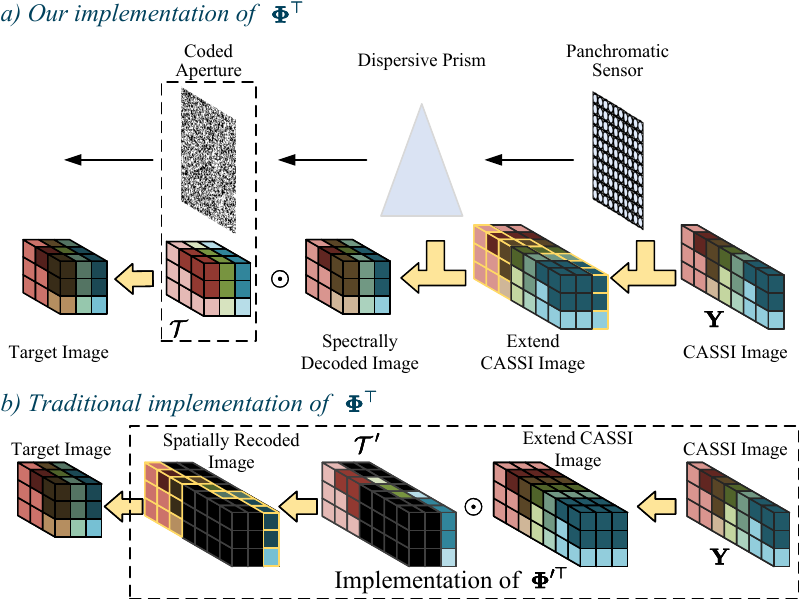}
    \caption{Visual comparison of the implementation of the adjoint operator. (a) Our implementation of the adjoint operator with complexity \(\mathcal{O}(HWC)\). (b) Traditional implementation of the adjoint operator with complexity \(\mathcal{O}(H(W+s(C-1))C)\).}
    \label{Visualization:PhiT}
\end{figure}

\begin{property}
The adjoint operator \(\mathbf{\Phi}^\top\) can be efficiently implemented with \(\mathcal{O}(HWC)\) complexity.  
\label{Property:Efficient_Adjoint}
\end{property}
\begin{proof}
Based on the definitions of \(\mathbf{\Phi}\) and \(\mathbf{\Phi}_c\), we have:
\begin{equation*}
    \mathbf{\Phi}^\top \boldsymbol{y} = \begin{bmatrix}
        \mathrm{diag}(\boldsymbol{t}_1)(\mathbf{I}_H \otimes \mathcal{K}_{:, :, 1})  \boldsymbol{y} \\
        \mathrm{diag}(\boldsymbol{t}_2)(\mathbf{I}_H \otimes \mathcal{K}_{:, :, 2}) \boldsymbol{y} \\
        \vdots \\
        \mathrm{diag}(\boldsymbol{t}_C)(\mathbf{I}_H \otimes \mathcal{K}_{:, :, C}) \boldsymbol{y} \\
    \end{bmatrix}.
\end{equation*}

Through inverse vectorization, the matrix form corresponding to \(\mathrm{diag}(\boldsymbol{t}_c)(\mathbf{I}_H \otimes \mathcal{K}_{:, :, c}) \boldsymbol{y}\) can be implemented as:
\begin{equation}
    \mathcal{T}_{:,:,c}\odot(\mathbf{Y}\mathcal{K}_{:,:,c}^\top) = \mathcal{T}_{:,:,c}\odot\mathbf{Y}_{:, s(c-1)+1:s(c-1)+W},
    \label{Equation:PhiT}
\end{equation}
which indicates that we only require \(\mathcal{O}(HWC)\) multiplications to obtain \(\mathbf{\Phi}^\top\boldsymbol{y}\). 
\end{proof}
{Indeed, \cref{Equation:PhiT} describes the reversed optical path from the image plane to the coded aperture plane, as shown in \Cref{Visualization:PhiT}.}

\begin{property} 
    \label{Property:PhiPhiT}
    The Gram matrix of \(\mathbf{\Phi}\), denoted as \(\mathbf{\Phi}\mathbf{\Phi}^\top\), is diagonal and identical to the Gram matrix of \(\mathbf{\Phi}'\) (i.e., \(\mathbf{\Phi}\mathbf{\Phi}^\top = \mathrm{diag}(\boldsymbol{\lambda}) = \mathbf{\Phi}'\mathbf{\Phi}'^\top\)).
\begin{proof}
        According to the definition of \(\mathbf{\Phi}\), we know \(\mathbf{\Phi}\mathbf{\Phi}^\top = \sum_{c=1}^C \mathbf{\Phi}_c \mathbf{\Phi}_c^\top\). Based on the definition of \(\mathbf{\Phi}_c\), we have:
\begin{equation*}
    \mathbf{\Phi}_c \mathbf{\Phi}_c^\top = \mathrm{blkdiag}
        ([\mathcal{K}_{:,:,c}^\top\mathrm{diag}(\mathcal{T}_{h,:,c})
        \mathcal{K}_{:,:,c}|\forall h]),
\end{equation*}
where \(\mathrm{blkdiag}(\cdot)\) denotes a block diagonal matrix consisting of the elements in the input list. It is easy to verify that \(\mathcal{K}_{:,:,c}^\top\mathrm{diag}(\mathcal{T}_{h,:,c})\mathcal{K}_{:,:,c}\) is a diagonal matrix. Thus, \(\mathbf{\Phi}_c\mathbf{\Phi}_c^\top\) is diagonal, and \(\mathbf{\Phi}\mathbf{\Phi}^\top\) is also diagonal. Therefore, we can express \(\mathbf{\Phi}\mathbf{\Phi}^\top = \mathrm{diag}(\boldsymbol{\lambda})\). Combining this with Property \ref{Property:Efficient_Adjoint}, we use \(\mathbf{\Phi}\mathbf{\Phi}^\top \boldsymbol{1}_{HW\times 1}\) to compute the matrix form of \(\boldsymbol{\lambda}\) as:
\begin{equation*}
    \mathbf{\Lambda} = \sum_{c=1}^{C} (\mathcal{T}_{:,:,c}\odot\mathcal{T}_{:,:,c})\mathcal{K}_{:,:, c}  = \sum_{c=1}^{C} (\mathcal{T}'_{:,:,c}\odot\mathcal{T}'_{:,:,c})
\end{equation*}
Hence, \(\mathbf{\Phi}\mathbf{\Phi}^\top = \mathrm{diag}(\boldsymbol{\lambda}) = \mathbf{\Phi}'\mathbf{\Phi}'^\top\). 
\end{proof}
\end{property}

{
\subsection{RGB Observation Model}
To complement the SD-CASSI measurements, the DC-CASSI system further provides an extra measurement via an auxiliary RGB camera. Here, we mathematically formulate the observation model of this auxiliary RGB camera, which retains critical spatial information otherwise lost during the spatial-spectral modulation and data compression processes inherent to the SD-CASSI system.

The observation model for the RGB measurement of a target scene is generally formulated as:
\begin{equation}
    \mathbf{Y}_\text{rgb} = \mathbf{X}\mathbf{T}_{\text{rgb}},
    \label{Equation:Observation_RGB}
\end{equation}
where \(\mathbf{Y}_\text{rgb}\in \mathbb{R}^{HW\times 3}\) denotes the matrix-form RGB image; \(\mathbf{T}_{\text{rgb}} \in \mathbb{R}^{C\times 3}\) represents the ideal spectral-to-RGB projection matrix, which maps the \(C\)-dimensional spectral information of spectral image \(\mathbf{X}\) to the three RGB color channels. From this formulation, it is clear that the RGB camera captures the spatial information of \(\mathbf{X}\) within a three dimensional subspace, which lays the groundwork for leveraging spatial priors to complement the spectral information from SD-CASSI in later reconstruction steps.}

{
\section{Adaptive Spatial Reference Image Generation}
\label{Init}
The RGB image provides high-fidelity spatial information, serving as a crucial prior for reconstruction. This section proposes an adaptive reference image generation strategy based on RGB image. It not only initializes the reference image for reconstruction in \Cref{sec:TVDS} but also updates it to match iteratively refined spectral results.

\subsection{Extracting Reference Image from RGB Subspace}
\label{sec:RGBsubspace}

For the matrix-form RGB image \(\mathbf{Y}_{\text{rgb}}\), we perform singular value decomposition to decouple spatial and color features: the left singular matrix \(\mathbf{U}_{\text{rgb}}\) of \(\mathbf{Y}_{\text{rgb}}\) explicitly encodes spatial information. Thus, a spectral image incorporating RGB spatial features is expressed as \(\mathbf{U}_{\text{rgb}}\mathbf{\Theta}\), where \(\mathbf{\Theta}\) denotes the unknown spectral coefficient matrix whose estimation requires auxiliary information. When the spectral image \(\mathbf{X}\) is available (e.g., intermediate reconstruction results), \(\mathbf{\Theta}\) is optimized by minimizing the Frobenius norm of the reconstruction residual:
\begin{equation}
    \underset{\mathbf{\Theta}}{\mathrm{argmin}} \quad \|\mathbf{X} - \mathbf{U}_{\text{rgb}}\mathbf{\Theta}\|_F^2.
\end{equation}  

The closed-form solution is \(\mathbf{\Theta} = \mathbf{U}_{\text{rgb}}^\top\mathbf{X}\), and the resulting reference image \(\mathbf{X}_\parallel = \mathbf{U}_{\text{rgb}}\mathbf{U}_{\text{rgb}}^\top\mathbf{X}\) corresponds to the projection of \(\mathbf{X}\) onto the RGB subspace. 

\Cref{Figure:SubspaceRGB} validates this representation: band-wise SSIM remains above 0.9 and PSNR exceeds 34~dB across all spectral bands, confirming that the RGB subspace retains high-fidelity spatial features for cross-sensor fusion.

\begin{figure}[htbp]
    \centering
    \adjustbox{padding=0pt}{
        \includegraphics[width=\linewidth]{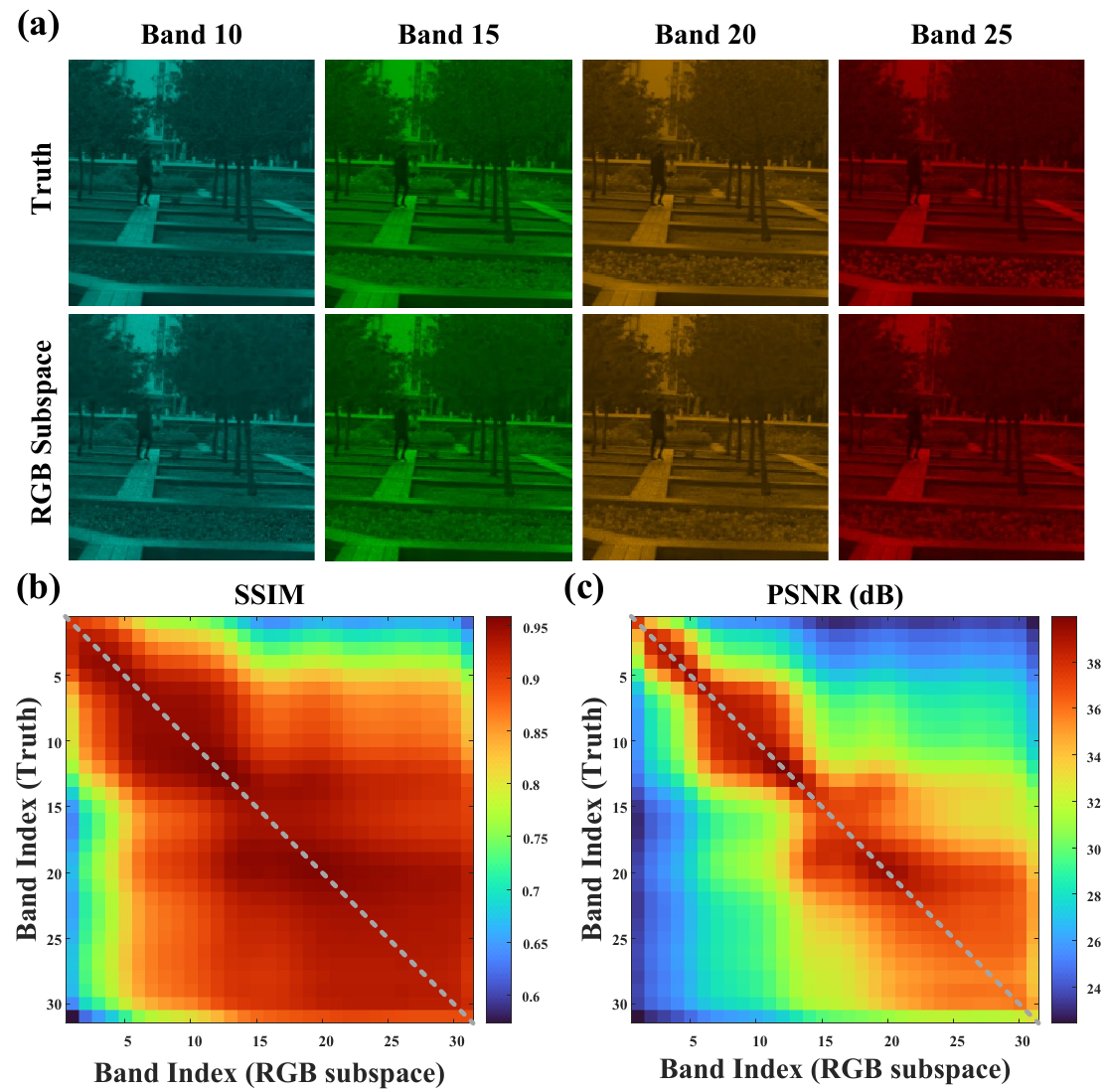}
    }
    \caption{Visualization of RGB subspace image of ARAD-0902 scene. (a) Visual comparison with ground truth in four bands. (b) Band-wise SSIM. (c) Band-wise PSNR.}
    \label{Figure:SubspaceRGB}
\end{figure}

{
However, the efficacy of this subspace fusion presupposes perfect pixel-wise correspondence between the RGB and CASSI sensors. In practical deployments, spatial misalignment between the two branches often renders the directly computed \(\mathbf{U}_{\text{rgb}}\) inaccurate. While hardware calibration~\cite{DC-CASSI_pan} can theoretically limit misalignment to negligible levels, practical factors such as device aging or mechanical shifts often introduce residual errors that compromise reconstruction fidelity. To bridge this gap between the ideal model and physical reality, we employ an Inception-style network optimized via self-supervised training to extract a refined \(\mathbf{U}_{\text{rgb}}\) in the presence of misalignment. Detailed architectural specifications and training protocols are provided in Supplementary Section~S1.
}

\subsection{Physics-Informed Initialization}
In practical scenarios, the spectral image \(\mathbf{X}\) is initially not given, making the direct projection onto the RGB subspace infeasible. Therefore, we integrate the physical observation model of the SD-CASSI to implement initialization. 

We formulate an optimization problem that balances two critical objectives: maintaining data fidelity to SD-CASSI measurements to preserve the reliability of the primary sensor's information and ensuring spatial consistency with the RGB subspace to leverage the auxiliary sensor's prior. This problem essentially forms an initial fusion task:
\begin{equation}
    \underset{\boldsymbol{x}, \mathbf{\Theta}}{\mathrm{argmin}}\quad \| \boldsymbol{y} - \mathbf{\Phi}\boldsymbol{x} \|^2 + \alpha \|\mathbf{X} - \mathbf{U}_{\text{rgb}}\mathbf{\Theta}\|_F^2, \label{Initial_Fusion}
\end{equation}
where \(\alpha\) adjusts the trade-off between the two terms, \(\boldsymbol{x}\) is the vectorized \(\mathbf{X}\). This formulation is consistent with the MAP estimation framework in \cref{fusionBayesian}. To solve this problem efficiently, we adopt the block coordinate descent method, which decomposes the joint optimization of $\boldsymbol{x}$ and $\mathbf{\Theta}$ into two sequential subproblems:
\begin{paragraph}{\textbf{Optimize \(\boldsymbol{x}\) with fixed \(\mathbf{\Theta}\)}}  
    The closed-form solution for the vectorized spectral image \(\boldsymbol{x}\) is:  
    {
   \begin{equation*}
       \boldsymbol{x}^* = \left(\alpha \mathbf{I} + \mathbf{\Phi}^\top\mathbf{\Phi}\right)^{-1}\left(\mathbf{\Phi}^\top\boldsymbol{y}+\alpha\mathrm{vec}(\mathbf{U}_{\text{rgb}}\mathbf{\Theta})\right).
   \end{equation*}  

   With Woodbury matrix identity to \(\left(\alpha \mathbf{I} + \mathbf{\Phi}^\top\mathbf{\Phi}\right)^{-1}\), we finally obtain the solution for \(\boldsymbol{x}\) as:
   \begin{equation}
   \boldsymbol{x}^* = \mathrm{vec}\left(\mathbf{U}_{\text{rgb}}\mathbf{\Theta}\right) + \mathbf{\Phi}^\top \left(\left(\boldsymbol{y} - \mathbf{\Phi}\mathrm{vec}\left(\mathbf{U}_{\text{rgb}}\mathbf{\Theta}\right)\right)\oslash (\boldsymbol{\lambda} + \alpha)\right), \label{Equation:X_init}
   \end{equation}}
   where \(\boldsymbol{\lambda}\) denotes the vector of diagonal elements of  \(\mathbf{\Phi}\mathbf{\Phi}^\top\) (see \Cref{Property:PhiPhiT}), \(\mathrm{vec}(\cdot)\) represents the vectorization operation, and \(\oslash\) denotes element-wise division.
\end{paragraph}
\begin{paragraph}{\textbf{Optimize \(\mathbf{\Theta}\) with updated \(\boldsymbol{x}^*\)}}
    Next, substitute \(\boldsymbol{x}^*\) into \cref{Initial_Fusion} and leverage Property \ref{Property:PhiPhiT} (\(\mathbf{\Phi}\mathbf{\Phi}^\top = \mathrm{diag}(\boldsymbol{\lambda})\)). This yields the following two equalities:  
\begin{align*}
    \boldsymbol{y} - \mathbf{\Phi}\boldsymbol{x}^* &= \alpha \cdot \mathrm{diag}(\alpha + \boldsymbol{\lambda})^{-1} \left( \boldsymbol{y} - \mathbf{\Phi}_{\mathbf{U}_{\text{rgb}}}\boldsymbol{\theta} \right), \\
    \boldsymbol{x}^* - \mathrm{vec}\left(\mathbf{U}_{\text{rgb}}\mathbf{\Theta}\right) &= \mathbf{\Phi}^\top \left( \mathrm{diag}(\alpha + \boldsymbol{\lambda})^{-1} \left( \boldsymbol{y} - \mathbf{\Phi}_{\mathbf{U}_{\text{rgb}}}\boldsymbol{\theta} \right) \right),
\end{align*}  
where \(\mathbf{\Phi}_{\mathbf{U}_{\text{rgb}}} = \mathbf{\Phi} \left( \mathbf{U}_{\text{rgb}} \otimes \mathbf{I}_{C} \right)\) and \(\boldsymbol{\theta}\) is the vectorized form of \(\mathbf{\Theta}\). Substituting the above equalities into the original objective function, we simplify it to a weighted least squares problem:  
\begin{align*}
    \min_{\boldsymbol{\theta}}\quad\mathrm{tr}\left( \left( \boldsymbol{y} - \mathbf{\Phi}_{\mathbf{U}_{\text{rgb}}}\boldsymbol{\theta} \right)^\top \mathrm{diag}(\alpha + \boldsymbol{\lambda})^{-1} \left( \boldsymbol{y} - \mathbf{\Phi}_{\mathbf{U}_{\text{rgb}}}\boldsymbol{\theta} \right) \right),
\end{align*}  
where \(\mathrm{tr}(\cdot)\) denotes the matrix trace operation. The solution is given by:
\begin{equation}
    \boldsymbol{\theta}^* = \left( \mathbf{\Phi}_{\mathbf{U}_{\text{rgb}}}^\top \mathrm{diag}(\alpha + \boldsymbol{\lambda})^{-1} \mathbf{\Phi}_{\mathbf{U}_{\text{rgb}}} \right)^{-1} \mathbf{\Phi}_{\mathbf{U}_{\text{rgb}}}^\top \mathrm{diag}(\alpha + \boldsymbol{\lambda})^{-1} \boldsymbol{y}.
\end{equation}
% To solve this problem using standard least-squares methods, we introduce two transformed variables:  
% \[
% \boldsymbol{y}' = \mathrm{diag}(\alpha + \boldsymbol{\lambda})^{-1/2} \boldsymbol{y}, \quad \mathbf{\Phi}_{\mathbf{U}_{\text{rgb}}}' = \mathrm{diag}(\alpha + \boldsymbol{\lambda})^{-1/2} \mathbf{\Phi}_{\mathbf{U}_{\text{rgb}}}.
% \] 

% Leveraging the proposed observation model, \(\mathbf{\Phi}_{\mathbf{U}_{\text{rgb}}}'\) can be computed efficiently by treating \(\mathbf{U}_{\text{rgb}} \otimes \mathbf{I}_{C}\) as a matrix with \(3C\) columns (each column corresponds to a vectorized spectral band image). The optimal solution \(\boldsymbol{\theta}^*\) is then obtained via the Moore-Penrose pseudo-inverse:  
% \begin{align*}
%     \boldsymbol{\theta}^* = \left( \mathbf{\Phi}_{\mathbf{U}_{\text{rgb}}}' \right)^\dagger \boldsymbol{y}'.
% \end{align*}
\end{paragraph}

Ultimately, we obtain the initial reference image \(\mathbf{X}_{\text{ref}} = \mathbf{U}_{\text{rgb}}\mathbf{\Theta}^*\) and the estimation of the spectral image \(\boldsymbol{x}^*\).}
% This physics-informed multi-source initial fusion not only ensures the reliability of the RGB sensor's spatial prior but also provides a robust initial condition for subsequent TV subgradient-guided reconstruction.}

\section{TV Subgradient-Guided Reconstruction}
\label{sec:TVDS}
{
\noindent Based on the generated reference image introduced in \Cref{Init}, the core of this section lies in {fusing} the structural information of the reference into spectral reconstruction through. We elaborate on four key aspects: the theoretical basis of subgradient-guided fusion, the design of the conditional prior, ADMM-based optimization, and algorithm analysis.
}
\subsection{Subgradient Guided Fusion}
To establish a rigorous theoretical basis for fusing CASSI spectral information and RGB spatial information, we first revisit the subgradient concept.
\begin{definition}[Subgradient]
    \label{Definition1}
    Let \( f: \mathbb{E} \to \bar{\mathbb{R}}\) be a convex function, with \( y \in \text{dom}\,f \). A vector \( s \in \mathbb{E} \) is called a subgradient of \( f \) at \( y \) if it satisfies the inequality:
    \begin{equation*}
        f(x) \geq f(y) + \langle s, x - y \rangle, \quad \forall x \in \text{dom}\,f,
    \end{equation*}
    where \( \langle \cdot, \cdot \rangle \) denotes the inner product. The set of all such subgradients at \( y \), called the subdifferential, is denoted \(\partial f(y)\). The function \( f \) is said to be subdifferentiable at \( y \) if \(\partial f(y)\neq \emptyset\). 
\end{definition}

Leveraging it, the following theorem can be derived:
\begin{theorem}
For any proper closed convex function \( f: \mathbb{E} \to \bar{\mathbb{R}}\) and a subgradient \( g \in \partial f(y) \) at \( y \in \mathrm{dom}f \), the solution set \( \mathcal{S} \triangleq \mathrm{arg\,min}_{x} \{ f(x) - \langle g, x \rangle \} \) satisfies that every \( x \in \mathcal{S} \) admits \( g \in \partial f(x) \). 
\end{theorem}
\begin{proof}
    With the optimization condition,  we have:
    \begin{equation*}
        f(z) \geq f(x) + \langle g, z - x\rangle, \quad \forall x \in \mathcal{S}, \forall z \in \mathrm{dom} f
    \end{equation*}
    According to the \Cref{Definition1}, \(\forall x \in \mathcal{S}\), the \(g\in\partial f(x)\). 
\end{proof}

Moreover, it is methodologically critical for image fusion: it guarantees that the subgradient features of reference image can be inherited by the subgradient-regularized result.

\subsection{Conditional Prior Design for Spatial Feature Fusion}
\label{TVDS_function}
Based on the subgradient theory, we select the TV function as the convex functional for the subgradient of TV describes local structural change directions, enabling the transfer of RGB spatial information to spectral reconstruction.

Formally, we define the 3D isotropic TV function for spectral images:
\begin{equation}
    R_{\text{TV}}(\mathcal{X}) = \sum_{h=1}^H \sum_{w=1}^W \sum_{c=1}^C \left\|(\Delta \mathcal{X})_{:, h,w,c}\right\|_{2},
    \label{Equation:TV_3D}
\end{equation}
where \(\Delta\) denotes the 2D spatial difference operator (computed per spectral band) with replicate padding:
\begin{align*}
    (\Delta\mathcal{X})_{1,h,w,:} =& \begin{cases} 
        \mathcal{X}_{h+1,w,:} - \mathcal{X}_{h,w,:}, & 1 \leq h < H \\
        \boldsymbol{0}_{C\times 1}, & \text{else} 
        \end{cases} \\
    (\Delta\mathcal{X})_{2, h,w,:} =& \begin{cases} 
        \mathcal{X}_{h,w+1,:} - \mathcal{X}_{h,w,:}, & 1 \leq w < W \\
        \boldsymbol{0}_{C\times 1}, & \text{else}.
        \end{cases}
\end{align*}
{where \((\Delta\mathcal{X})_{1,h,w,:}\) and \((\Delta\mathcal{X})_{2, h,w,:}\) represent vertical and horizontal spatial differences at pixel \( (h,w) \), respectively.}

{
It is straightforward to show that the subgradient of the TV function at \(\mathcal{X}\) is given by \(-\mathrm{div} \mathcal{P}_{\mathcal{X}}\), where:}
\begin{itemize}
    \item \(\mathcal{P}_{\mathcal{X}}\) is the normalized spatial difference of \(\mathcal{X}\), explicitly encoding the direction of local structural changes. Specifically, it's defined as:
    \begin{equation}
        \mathcal{P}_{\mathcal{X}} = (\Delta \mathcal{X}) \oslash { \|\Delta \mathcal{X} \|_{2,\text{dim}=1}}, 
        \label{Equation:Definition_P}
    \end{equation}
where $\|\cdot\|_{2,\text{dim}=1}$ denotes the $\ell_{2}$-norm computed along the first dimension of \(\Delta \mathcal{X}\). {However, this definition becomes singular where \((\Delta \mathcal{X})_{:,h,w,c} = 0\). In such cases, \((\mathcal{P}_{\mathcal{X}})_{:,h,w,c}\) lies in the unit ball~\cite{TV_Chambolle}. Thus, \(\mathcal{P}_{\mathcal{X}}\) is non-unique, and all admissible solutions form the set \(\mathbb{S}_{\mathcal{X}}\).}
\item 
\(\mathrm{div}\) is the discrete divergence operator corresponding to \(\Delta\), which converts local difference directions into structural consistency, defined as
\begin{align*}
    (\mathrm{div}\mathcal{P}_{\mathcal{X}})_{h,w,:} =& (\mathcal{P}_{\mathcal{X}})_{1,h,w,:} - (\mathcal{P}_{\mathcal{X}})_{1,h-1,w,:} \\ &+ (\mathcal{P}_{\mathcal{X}})_{2,h,w,:} - (\mathcal{P}_{\mathcal{X}})_{2,h,w-1,:},
\end{align*}
\end{itemize}

Notably, \(\mathrm{div}\mathcal{P}_{\mathcal{X}}\) describes the consistency of local structural change directions in horizontal and vertical dimensions, acting as robust spatial features. 
% For flat image regions where \(\|(\Delta \mathcal{X})_{:, h,w,c}\|_2=0\), \(\mathcal{P}_{\mathcal{X}}\) and \(\mathrm{div}\mathcal{P}_{\mathcal{X}}\) become undefined, so we use \Cref{Algorithm:TVDS_Fusion} to compute the subgradient.

To achieve spatial information fusion, we construct the TVDS regularization term by incorporating the TV subgradient of the RGB-derived reference image \(\mathcal{X}_\text{ref}\).
\begin{equation}
    R_\text{TVDS}(\mathcal{X};\mathcal{X}_\text{ref}) = R_\text{TV}(\mathcal{X}) + \langle \mathrm{div}\mathcal{P}_{\mathcal{X}_\text{ref}}, \mathcal{X} \rangle,
    \label{Equation:TVDS_Constraint}
\end{equation}
{where \(-\mathrm{div}\mathcal{P}_{\mathcal{X}_\text{ref}}\) is a subgradient of the TV function at the point \(\mathcal{X}_\text{ref}\).}
It is noteworthy that if \(\mathrm{div}\mathcal{P}_{\mathcal{X}_\text{ref}}\) is set to zero (i.e., no reference guidance), \(R_\text{TVDS}\) degrades to the conventional TV regularization term. Thus, TVDS can be interpreted as an extension of TV for guided spatial information fusion.
\subsection{ADMM-Based Reconstruction with TVDS}
Based on the TVDS term, we formulate the DC-CASSI reconstruction problem as \cref{fusionBayesian}:
\begin{equation}
    \underset{\mathcal{X}}{\text{argmin}} \quad \frac{1}{2} \Vert \boldsymbol{y} - \mathbf{\Phi} \boldsymbol{x} \Vert^2 + \mu R_{\text{TVDS}}(\mathcal{X};\mathcal{X}_\text{ref}),
    \label{Problem:DC_CASSI_TVDS}
\end{equation}
where \(\mu>0\) controls the trade-off between data fidelity and regularization strength of the TVDS term.
Directly solving \cref{Problem:DC_CASSI_TVDS} is challenging due to the non-smooth nature of \(R_{\text{TVDS}}\). {We} address this via an auxiliary variable \(\mathcal{Z}\) and reformulate the problem as a constrained optimization:
\begin{align}
    \underset{\mathcal{X}, \mathcal{Z}}{\arg\min} &\quad \frac{1}{2}\| \boldsymbol{y} - \mathbf{\Phi}\boldsymbol{x} \|_2^2 + \mu R_{\text{TVDS}}(\mathcal{Z};\mathcal{X}_\text{ref}) \notag\\
    \text{s.t.} &\quad \mathcal{X} = \mathcal{Z}.
    \label{Equation:Auxiliary}
\end{align}

Via the augmented Lagrangian method, we transform \cref{Equation:Auxiliary} into an unconstrained one, which is then solved via the scaled alternating direction method of multipliers (ADMM):
\begin{align}
    \mathcal{X}^{(n+1)} &= \underset{\mathcal{X}}{\arg\min} \left\{\| \boldsymbol{y} - \mathbf{\Phi}\boldsymbol{x} \|_2^2 + \rho \| \mathcal{X} - \mathcal{Z}^{(n)} + \mathcal{U}^{(n)}\|_F^2\right\}, \label{ADMM_step1} \\
    \mathcal{Z}^{(n+1)} &= \underset{\mathcal{Z}}{\arg\min}  \Big\{\frac{1}{2}\| \mathcal{Z} - \mathcal{X}^{(n+1)} - \mathcal{U}^{(n)}\|_F^2 \notag \\ & \qquad\qquad\qquad\qquad + \eta R_{\text{TVDS}}(\mathcal{Z};\mathcal{X}_\text{ref})\Big\}, \label{ADMM_step2}\\
    \mathcal{U}^{(n+1)} &= \mathcal{U}^{(n)} + \mathcal{X}^{(n+1)} - \mathcal{Z}^{(n+1)},\label{ADMM_step3}
\end{align}
where \(\rho>0\) is the penalty parameter, \(n\) is the ADMM iteration index, \(\mathcal{U}\) is the scaled Lagrange multiplier, and \(\eta\triangleq\mu\rho^{-1}\).  

{By vectorizing \cref{ADMM_step1} (where \(\boldsymbol{x}, \boldsymbol{z}, \boldsymbol{u}\) denote the vectorized counterparts of \(\mathcal{X}, \mathcal{Z}, \mathcal{U}\)) and applying the Woodbury matrix identity, we obtain the closed-form solution:}
\begin{equation}
    \begin{aligned}
        &\boldsymbol{x}^{(n + 1)} = \underset{\boldsymbol{x}}{\arg\min} \| \boldsymbol{y} - \mathbf{\Phi}\boldsymbol{x} \|_2^2 + \rho \| \boldsymbol{x} - \boldsymbol{z}^{(n)} + \boldsymbol{u}^{(n)} \|_2^2  \\ 
    &= \boldsymbol{z}^{(n)} - \boldsymbol{u}^{(n)} + \mathbf{\Phi}^{\top} \left( \left( \boldsymbol{y} - \mathbf{\Phi} (\boldsymbol{z}^{(n)} - \boldsymbol{u}^{(n)}) \right) \oslash (\rho + \boldsymbol{\lambda}) \right), 
    \end{aligned}
    \label{Equation:ADMM_step1_vec_solution}
\end{equation}
where \(\boldsymbol{\lambda}\) denotes the diagonal of \(\mathbf{\Phi}\mathbf{\Phi}^{\top}\) (see \Cref{Property:PhiPhiT}). This step can be computed with a computational complexity of $\mathcal{O}(HWC)$ by leveraging \Cref{Property:Efficient_Adjoint}.

For \cref{ADMM_step2}, let \(\mathcal{Z}_{\text{temp}} = \mathcal{X}^{(n+1)} + \mathcal{U}^{(n)}\). 
We can derived the equation from optimality condition:
% The optimality condition is derived as:
{
% \begin{equation*}
%     \mathcal{Z} - \mathcal{Z}_{\text{temp}} +\eta\mathrm{div}\mathcal{P}_{\mathcal{X}_\text{ref}} \in \{\eta\mathrm{div}\mathcal{P}_{\mathcal{Z}}|\mathcal{P}_{\mathcal{Z}}\in\mathbb{S}_{\mathcal{Z}}\},
% \end{equation*}
% which implies that the optimal solution \(\mathcal{Z}^*\) and a chosen element \(\mathcal{P}_{\mathcal{Z}^*}^* \in \mathbb{S}_{\mathcal{Z}^*}\) satisfy the following fixed-point equation:
\begin{equation}
    \mathcal{Z}^* = \mathcal{Z}_{\text{temp}} -\eta\mathrm{div}\mathcal{P}_{\mathcal{X}_\text{ref}} + \eta\mathrm{div}\mathcal{P}_{\mathcal{Z}^*}^*, \label{Equation:FixedPoint_Z}
\end{equation}
where \(\mathcal{Z}^*\) is the optimal solution and \(\mathcal{P}_{\mathcal{Z}^*}^* \in \mathbb{S}_{\mathcal{Z}^*}\).

To implement \cref{Equation:FixedPoint_Z} in practice, we address the singularity issue in the original definition of $\mathcal{P}_{\mathcal{Z}}$ by introducing a step-size parameter $\tau > 0$ and reformulating \cref{Equation:Definition_P} as follows:
\begin{align}
    % &\mathcal{P}_{\mathcal{Z}} = (\Delta \mathcal{Z}) \oslash { \|\Delta \mathcal{Z} \|_{2,\text{dim}=1}} \notag\\ 
    % &\implies \tau\mathcal{P}_{\mathcal{Z}} \odot {\|\Delta \mathcal{Z} \|_{2,\text{dim}=1}} = \tau\Delta \mathcal{Z} \notag \\
    % &\implies \mathcal{P}_{\mathcal{Z}} \left(1 + \tau\|\Delta \mathcal{Z} \|_{2,\text{dim}=1}\right) = \mathcal{P}_{\mathcal{Z}} + \tau\Delta \mathcal{Z} \notag \\ &\implies
    \mathcal{P}_{\mathcal{Z}} = \left(\mathcal{P}_{\mathcal{Z}} + \tau\Delta \mathcal{Z}\right) \oslash \left(1 + \tau\|\Delta \mathcal{Z} \|_{2,\text{dim}=1}\right).
    \label{Equation:P_update}
\end{align}

By substituting \cref{Equation:P_update} into \cref{Equation:FixedPoint_Z} and denoting iterative updates with superscript $[k]$, we obtain the following coupled fixed-point iteration scheme:}
\begin{align}
    \mathcal{Z}^{[k+1]} =& \mathcal{Z}_{\text{temp}} -\eta\text{div}\mathcal{P}_{\mathcal{X}_{\text{ref}}} +\eta\text{div}\mathcal{P}_{\mathcal{Z}}^{[k]}, \label{FPI1}\\
    \mathcal{P}_{\mathcal{Z}}^{[k+1]} =& \left(\mathcal{P}_{\mathcal{Z}}^{[k]} + \tau\Delta \mathcal{Z}^{[k+1]}\right) \oslash \left(1 + \tau\|\Delta \mathcal{Z}^{[k+1]} \|_{2,\text{dim}=1}\right), \label{FPI2}
\end{align}
where \(k=1,2,\dots,K\) denotes the fixed-point iteration index. We initialize \(\mathcal{P}_{\mathcal{Z}}^{[0]}=\boldsymbol{0}\) and set \(\tau=1/(8\eta)\) to satisfy the convergence condition~\cite{TV_Chambolle}. The fixed-point iteration procedure is summarized in \Cref{Algorithm:TVDS_Fusion}.

\begin{algorithm}[ht]
    \caption{TV Subgradient Guided Fusion}
    \KwIn{Initial Spectral image \(\mathcal{Z}_{\text{temp}}\), reference subgradient \(-\mathrm{div}\mathcal{P}_{\mathcal{X}_\text{ref}}\), regularization parameter \(\eta\) and maximum iterations  \(K\).}
    Set \(k\gets 0, \mathcal{P}^{[0]} \gets\boldsymbol{0}, \tau\gets 1/(8\eta)\);\\
    \While{\(k<K\)}{
        Update \(\mathcal{Z}^{[k+1]}\) by \cref{FPI1};\\
        Update \(\mathcal{P}_{\mathcal{Z}}^{[k+1]}\) by \cref{FPI2};\\
        \(k\gets k+1\);
    }
    \KwOut{Fused image \(\mathcal{Z}^{[K]}\), estimated subgradient \(-\mathrm{div}\mathcal{P}_{\mathcal{Z}}^{[K]}\)}
    \label{Algorithm:TVDS_Fusion}
\end{algorithm}

\subsection{Discussion on the ADMM Solution}
To verify the reliability of the proposed algorithm, we analyze its convergence and present the full reconstruction pipeline with stage-wise parameter updates.

\begin{theorem}
    The ADMM solution converges as \(\rho\) increases. \label{Theorem:ADMM_convergence} 
\end{theorem}
\begin{proof}
    From Step 2 of ADMM-TVDS and the inequality \(\|\mathrm{div}\mathcal{P}_{\mathcal{Z}}\|_F\leq \sqrt{8}\|\mathcal{P}_{\mathcal{Z}}\|_F\) (Proven in \cite{TV_Chambolle}), we derive the bound for \(\|\mathcal{U}^{(n+1)}\|_F\):
    \begin{align*}
        \|\mathcal{U}^{(n+1)}\| &=\|\mathcal{Z}^{(n+1)} - \mathcal{X}^{(n+1)} - \mathcal{U}^{(n)}\|_F \\
         &= \eta\|\mathrm{div}\mathcal{P}_{\mathcal{Z}}^{[K]} - \mathrm{div}\mathcal{P}_{\mathcal{X}_{\text{ref}}}\|_F\\
&\leq\sqrt{8}\mu\rho^{-1} \|\mathcal{P}_{\mathcal{Z}}^{[K]} - \mathcal{P}_{\mathcal{X}_{\text{ref}}}\|_F \\
&\leq 8\mu\rho^{-1}\sqrt{HWC}:=L_1(\rho)
    \end{align*}
    From Step 1 of ADMM-TVDS, we have
\begin{align*}
    &\| \boldsymbol{x}^{(n + 1)} - \boldsymbol{z}^{(n)} + \boldsymbol{u}^{(n)}\| \\
    &\quad= \left\|\mathbf{\Phi}^{\top} (( \boldsymbol{y} - \mathbf{\Phi} (\boldsymbol{z}^{(n)} - \boldsymbol{u}^{(n)})) \oslash (\rho + \boldsymbol{\lambda}) )\right\| \\
    &\quad\leq \frac{\sqrt{\lambda_{\max}}\|\boldsymbol{y}\| + \lambda_{\max} \|\boldsymbol{z}^{(n)}-\boldsymbol{u}^{(n)}\|}{\rho+\lambda_{\min}}\\
    &\quad\leq \frac{\sqrt{\lambda_{\max}}\|\boldsymbol{y}\| + \lambda_{\max}L_1(\rho)+\lambda_{\max} \sqrt{HWC}}{\rho+\lambda_{\min}}:=L_2(\rho)
\end{align*}

where \(\lambda_{\max}\) and \(\lambda_{\min}\) are the maximum and minimum values of \(\mathbf{\Phi}\mathbf{\Phi}^{\top}\), respectively. By leveraging \cref{ADMM_step3}, the iteration errors can be derived:
    \begin{align*}
        \|\mathcal{U}^{(n+1)} - \mathcal{U}^{(n)} \|_F & \leq \|\mathcal{U}^{(n+1)}\|_F + \|\mathcal{U}^{(n)} \|_F \leq 2L_1(\rho) \\
        \|\mathcal{Z}^{(n+1)} - \mathcal{Z}^{(n)} \|_F &= \|\mathcal{X}^{(n+1)}-\mathcal{Z}^{(n)}+\mathcal{U}^{(n)}-\mathcal{U}^{(n+1)}\|_F \\ &\leq L_1(\rho)+L_2(\rho) \\
        \|\mathcal{X}^{(n+1)} - \mathcal{X}^{(n)} \|_F &= \|\mathcal{X}^{(n+1)}-\mathcal{Z}^{(n)}-\mathcal{U}^{(n)}+\mathcal{U}^{(n-1)}\|_F \\
        &\leq 3L_1(\rho) + L_2(\rho)
    \end{align*}

    As \(\rho\) increases, both \(L_1(\rho)\) and \(L_2(\rho)\) both decrease, Consequently, all iteration errors tend to zero as \(\rho\) grows without bound. This completes the proof of the convergence of the ADMM-TVDS algorithm. 
\end{proof}
\begin{algorithm}[htbp]
    \caption{ADMM-TVDS for DC-CASSI}
    \KwIn{SD-CASSI sensing matrix \(\mathbf{\Phi}\), measurement \(\boldsymbol{y}\), RGB image \(\mathcal{Y}_{\text{rgb}}\), parameters \(\alpha\), \(\mu\), \(\rho\), \(\beta\),\(L\), \(N\), \(K\).}
    Initialize \(\mathbf{U}_{\text{rgb}}\), \(\mathcal{X}^{(N)}\), \(\mathcal{X}_{\text{ref}}\) as \Cref{Init}; \\
    Set \(\mathcal{U}^{(N)}\gets \boldsymbol{0}\), stage index \(l\gets0\); \\
    \While{\(l<L\)}{
        Compute \(-\mathrm{div}\mathcal{P}_{\mathcal{X}_\text{ref}}\) with spectral image $\mathcal{X}_{\text{ref}}$ and reference subgradient $\mathbf{0}$ as inputs;\\
        Set \(\mathcal{U}^{(0)}\gets\mathcal{U}^{(N)}\), \(\mathcal{X}^{(0)}\gets\mathcal{X}^{(N)}\), \(n\gets 0\); \\
        \While{\(n<N\)}{
            Update \(\mathcal{Z}^{(n+1)}\) via \Cref{Algorithm:TVDS_Fusion};\\
            Update \(\mathcal{U}^{(n+1)}\) via \cref{ADMM_step3};\\
            Update \(\mathcal{X}^{(n+1)}\) via \cref{Equation:ADMM_step1_vec_solution};\\
            \(n\gets n +1\)
        }
        Update reference image:\(\mathbf{X}_{\text{ref}}\gets \mathbf{U}_{\text{rgb}}\mathbf{U}_{\text{rgb}}^\top \mathbf{X}^{(N)}\); \\
        Increase penalty parameter:\(\rho \gets \beta\rho \);\\
        \(l\gets l+1\)\\
    }
     \KwOut{Reconstructed spectral image 
 \(\mathcal{X}^{(N)}\).}
    \label{ADMM-TVDS}
\end{algorithm}
Based on Theorem \ref{Theorem:ADMM_convergence}, we design a stage-wise  algorithm (\Cref{ADMM-TVDS}) to enhance reconstruction performance. In each stage, the penalty parameter \(\rho\) (via a growth factor \(\beta\)) and the reference subgradient \(-\mathrm{div}\mathcal{P}_{\mathcal{X}_\text{ref}}\)  are updated.

Neglecting trivial operations (addition, assignment), the computational complexity of the key algorithmic steps is:
\begin{itemize}
    \item Update \(\mathcal{X}^{(n+1)}\): \(\mathcal{O}(HWC)\);
    \item Update \(\mathcal{Z}^{(n+1)}\): \(\mathcal{O}\left(HWCK\right)\);
    \item Update reference image: \(\mathcal{O}\left(HWC\right)\);
\end{itemize}

The overall complexity is \(\mathcal{O}\left(HWCKNL\right)\), this linear scaling with image size ensures the algorithm is feasible for high-resolution spectral images.

\section{Experiments on Simulated Datasets}
\label{sec6}
To fully validate the proposed reconstruction method, we conduct experiments on simulated datasets in this section. Subsequent parts are organized as follows: first introducing datasets and experimental setup, then presenting benchmark comparative results, followed by noise robustness analysis, ablation studies, and computational complexity analysis. These ensure a comprehensive evaluation of TVDS's performance.
\subsection{Datasets description}
We use public ideal datasets for comparative and generate noisy datasets for robustness analysis.
\paragraph{\textbf{Ideal Datasets}}
Consistent with previous DC-CASSI studies \cite{ADMM_PIDS, In2SET_2024, PiE, MLP-AMDC_2025, CasFormer}, two spectral image datasets, TSA-Simu and CAVE-28, are employed to evaluate the performance under ideal observation scenarios. Specifically, TSA-Simu (\(256\times256\times28\)) is generated from the KAIST dataset \cite{KAIST_dataset}, with its coded aperture mask obtained via real-world measurements as reported in \cite{TSA_Net}. To construct the DC-CASSI version of the dataset, we adopt RGB data sourced from \cite{MLP-AMDC_2025}, which corresponds to an ideal projection integrating the spectral response function (SRF) of a real RGB camera.
For the CAVE-28 dataset (\(512\times512\times28\)), we maintain alignment with CasFormer \cite{CasFormer}, facilitating fair comparison of multi-source fusion performance.

\begin{figure}[ht]
    \centering
    \includegraphics[width=\linewidth]{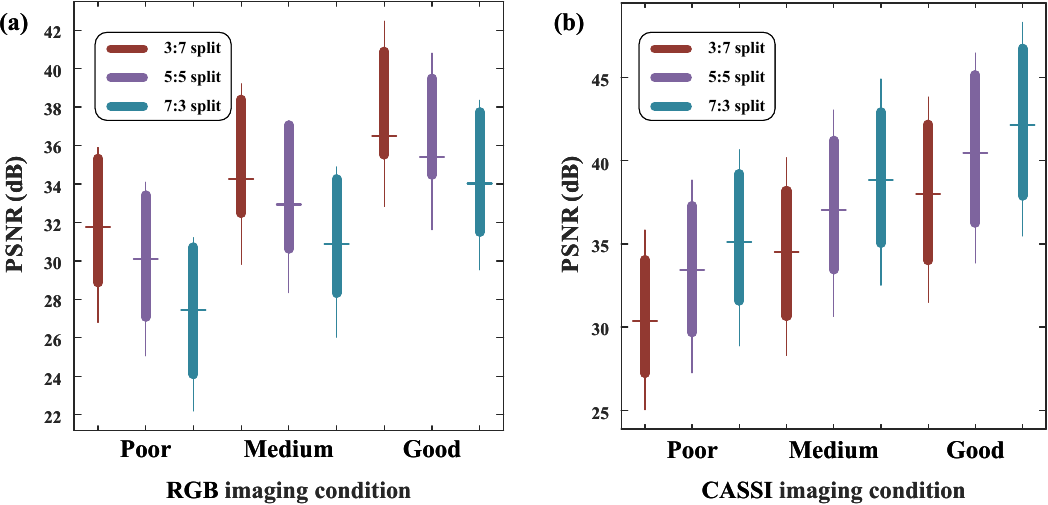}
    \caption{Statistical PSNR of generated RGB/SD-CASSI images of the ARAD-Simu dataset, {with colors indicating distinct light split ratios (CASSI:RGB)}. (a) RGB images across imaging conditions; (b) SD-CASSI images under identical settings.}
    \label{Fig:Noisy_Dataset}
\end{figure}

\paragraph{\textbf{Noisy Dataset}} 
\begin{figure*}[htbp]
    \centering\includegraphics[width=0.78\linewidth]{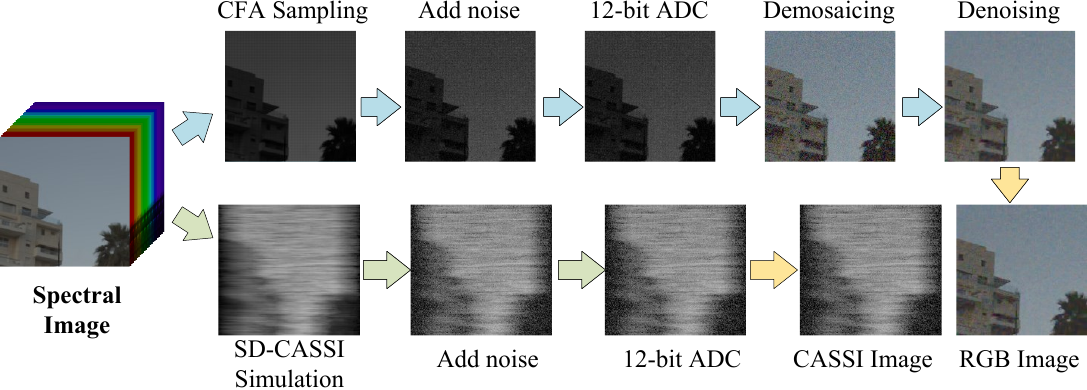}\caption{Simulation pipeline of the ARAD-Simu dataset}\label{Fig:ARAD_Simu}\end{figure*}
To evaluate noise robustness, we constructed the ARAD-Simu dataset (code available\footnote{https://github.com/bestwishes43/ADMM-TVDS}) following the pipeline in \Cref{Fig:ARAD_Simu}, addressing the absence of dedicated noisy DC-CASSI benchmarks. Derived from the ARAD dataset \cite{NTIRE2022}, we center-cropped spectral images to \(256\times256\times31\) for compatibility with TSA-Simu coded aperture mask and generated paired RGB and SD-CASSI observations. RGB images were produced using official NTIRE2022 tools \cite{NTIRE2022} with traditional demosaicing \cite{Demosaic_Menon_2007}, while CASSI measurements followed \cref{Equation:Former_Observation}.
We modeled realistic imaging degradations using Gaussian and Poisson noise, whose levels were controlled by ISO, exposure time, {and the light split ratio between the CASSI and RGB.} Three imaging conditions were defined: \textit{Poor} (short exposure + high ISO), \textit{Moderate} (default parameters), and \textit{Good} (long exposure + low ISO). Combined with three light split ratios (3:7, 5:5, 7:3), this yielded nine noisy configurations summarized in \Cref{Fig:Noisy_Dataset}.

\subsection{Experimental Setup}
\paragraph{\textbf{Metrics}}
We employ three complementary metrics: PSNR (dB) for reconstruction fidelity, SSIM for structural similarity, and SAM (°) for spectral consistency. This combination ensures comprehensive evaluation of spatial and spectral reconstruction quality. For consistency, bold values in all tables denote the optimal performance.
\paragraph{\textbf{Compared Methods}}
We evaluate TVDS against state-of-the-art DC-CASSI reconstruction methods:
\begin{itemize}
\item \textbf{Model-based}: \textbf{PIDS}~\cite{ADMM_PIDS} constrains reconstruction using spatial difference from RGB-derived prior image.
\item \textbf{Self-supervised}: \textbf{PiE}~\cite{PiE} fuses PIDS reconstructions with autoencoder outputs; \textbf{SIGDU-Net}~\cite{SIGDU-Net} employs self-supervised deep unfolding with spatial-spectral attention, { and trained with spatial and spectral TV loss}.
\item \textbf{Supervised}: \textbf{In2SET}~\cite{In2SET_2024} integrates sensor quantum efficiency priors with attention mechanisms; \textbf{CasFormer}~\cite{CasFormer} uses cascaded Transformers for spatial-spectral fusion.
\end{itemize}

Due to varying code/data availability, comparisons were adjusted for fairness: SIGDU-Net results are reported only on CAVE-28 using published data; CasFormer is evaluated on CAVE-28 with provided weights; and In2SET, following standard practice for supervised CASSI reconstruction methods, was trained on CAVE-28 and evaluated on TSA-Simu.

\paragraph{\textbf{Implementation Details}}

{
The proposed TVDS method is implemented with parameters detailed in \Cref{Tab:setting}.

\begin{table}[ht]
    \centering
    \caption{Parameter configuration of Proposed method.}
    \resizebox{0.95\linewidth}{!}{
    \begin{tabular}{ccc}
    \toprule
    \textbf{Parameters} & \textbf{Noise-free case} & \textbf{Noisy case} \\
    \midrule
    \(\rho\)     & 0.03  & 30 \\
    \(\mu\) & 0.03 & 30 \\
    \(\alpha\) & \multicolumn{2}{c}{0.1} \\
    \(\beta\) & \multicolumn{2}{c}{1.2} \\
    Number of Stages (\(L\)) & \multicolumn{2}{c}{30}\\ 
    ADMM iterations (\(N\))  & \multicolumn{2}{c}{10} \\
    Fixed point iterations (\(K\)) & \multicolumn{2}{c}{30} \\
    \bottomrule
    \end{tabular}
    }
    \label{Tab:setting}
\end{table}

For noise-free scenarios, low values of $\mu$ and $\rho$ (0.03) prioritize the accurate SD-CASSI model constraint. A consistent parameter set is maintained across all datasets to demonstrate system compatibility despite varying sensor configurations. In noisy conditions, elevated $\mu$ and $\rho$ values (30) enhance multi-source fusion regularization to suppress noise propagation. A single parameter configuration is applied universally across all noise levels to validate noise robustness and parameter robustness. All experiments were conducted on an NVIDIA RTX 5070Ti GPU platform.
}
\subsection{Comparison on CAVE-28 dataset}

As summarized in Table \ref{Tab:CAVE_28}, TVDS achieves the best performance across all three core metrics, demonstrating its comprehensive reconstruction capability. 

\begin{table}[htbp]
  \centering
    \caption{Quantitative evaluation on the CAVE-28 dataset}
    \resizebox{\linewidth}{!}{
    \begin{tabular}{cccc}
    \toprule[1pt]
    \multicolumn{1}{c}{Method} & PSNR  & SSIM  & SAM \\
    \midrule
    
    \makecell{PIDS (2023, TPAMI)} & 37.54±3.6 & 0.975±0.01 & 4.727±1.73 \\
    \makecell{CasFormer (2024, INFFUS)} & 37.24±3.2 & 0.983±0.01 & 4.648±1.31 \\
    \makecell{PiE (2024, TCSVT)} & 46.19±1.8 & 0.994±0 & 1.862±0.48 \\
    \makecell{SIGDU-Net (2025, INFFUS)} & 45.95±4.5 & 0.988±0.01 & 3.29±3.37 \\
    \makecell{\textbf{Initialization (Ours)}} & 34.73±4.5 & 0.778±0.16 & 9.253±4.6 \\
    \makecell{ \textbf{TVDS (Ours)}} & \textbf{48.29}±4.4 & \textbf{0.996}±0 & \textbf{1.363}±0.54 \\
    \bottomrule[1pt]
\end{tabular}
    }
\label{Tab:CAVE_28}
\end{table}

In terms of average PSNR, TVDS reaches 48.29 dB, outperforming the second-ranked PiE by approximately 2.1 dB. For average SSIM, TVDS attains 0.996, which is close to 1, reflecting its exceptional ability to preserve the spatial structural details of the original spectral image. In the average SAM metric, TVDS achieves the lowest value of 1.363°, outperforming PiE (1.862°) and SIGDU-Net (3.29°), confirming its superior preservation of critical spectral features.

\begin{table*}[!ht]
    \centering
    \setlength{\tabcolsep}{4 pt}
    \caption{Quantitative evaluation on the TSA-simu dataset}
    \resizebox{\linewidth}{!}{
    \begin{tabular}{ccccccccccccc}
    \toprule[1pt]
    Method & Metrics & S01   & S02   & S03   & S04   & S05   & S06   & S07   & S08   & S09   & S10   & Avg    \\
    \midrule
    \multirow{3}[2]{*}{\makecell{Initialization \\ (Ours)}}
&PSNR  & 40.86  & 39.34  & 35.74  & 44.02  & 41.97  & 42.28  & 44.10  & 41.76  & 33.37  & 45.81  & 40.92  \\
    &SSIM  & 0.971  & 0.962  & 0.926  & 0.971  & 0.974  & 0.952  & 0.988  & 0.954  & 0.875  & 0.965  & 0.954  \\
    &SAM   & 4.884  & 7.210  & 6.031  & 10.332  & 4.533  & 7.970  & 3.066  & 9.364  & 12.264  & 7.969  & 7.362  \\

     \midrule
     \multirow{3}[2]{*}{\makecell{PIDS \\ (2023, TPAMI)}} 
     & PSNR & 37.01 & 36.56 & 34.38 & 45.45 & 35.05 & 33.91 & 33.47 & 35.06 & 34.71 & 38.08 & 36.37  \\
     & {SSIM} & {0.97} & {0.919} & {0.93} & {0.97} & {0.945} & {0.934} & {0.92} & {0.935} & {0.942} & {0.959} & {0.942}  \\
    
     & SAM  & 8.374 & 18.172 & 10.927 & 12.169 & 11.173 & 17.055 & 11.368 & 18.266 & 10.167 & 14.305 & 13.198  \\

     \midrule
     \multirow{3}[2]{*}{\makecell{In2SET-9stg\\(2024, CVPR)}} 
& PSNR  & 42.59 & 46.45 & \textbf{44.58} & \textbf{50.67} & 42.04 & 42.53 & 41.62 & 40.57 & 43.86 & 42.36 & 43.73 \\
          & SSIM  & 0.987 & 0.992 & 0.983 & 0.992 & 0.99  & 0.988 & 0.981 & 0.986 & 0.987 & 0.991 & 0.988 \\
          & SAM   & 4.932 & 6.674 & 3.89  & 8.592 & 4.234 & 8.641 & 4.041 & 9.386 & 5.357 & 7.736 & 6.348 \\
    \midrule
     \multirow{3}[2]{*}{\makecell{PiE \\ (2024, TCSVT)}} 
          & PSNR  & 44.13 & 44.17 & 42.41 & 48.08 & 42.99 & 44.1  & 44.72 & 43.56 & 42.46 & 46.24 & 44.28 \\
          & SSIM  & 0.992 & 0.991 & \textbf{0.984} & 0.995 & 0.991 & 0.992 & 0.989 & 0.993 & 0.984 & 0.997 & 0.991 \\
          & SAM   & 3.251 & 5.292 & \textbf{2.843} & 4.658 & 3.626 & 4.795 & 3.069 & 4.545 & 5.137 & 4.59  & 4.181 \\
    \midrule
     \multirow{3}[2]{*}{\makecell{TVDS \\ (Ours)}} 
& PSNR  & \textbf{45.75} & \textbf{50.98} & 42.9  & 50.44 & \textbf{45.94} & \textbf{46.26} & \textbf{45.56} & \textbf{46.4} & \textbf{46.06} & \textbf{49.09} & \textbf{46.94} \\
          & SSIM  & \textbf{0.995} & \textbf{0.997} & \textbf{0.984} & \textbf{0.997} & \textbf{0.996} & \textbf{0.996} & \textbf{0.991} & \textbf{0.995} & \textbf{0.991} & \textbf{0.999} & \textbf{0.994} \\
          & SAM   & \textbf{2.576} & \textbf{3.98} & 3.158 & \textbf{2.843} & \textbf{1.573} & \textbf{3.123} & \textbf{2.951} & \textbf{4.352} & \textbf{3.675} & \textbf{1.727} & \textbf{2.996} \\

    \bottomrule
    \end{tabular}%
    }
    \label{Tab:TSA-simu}
    \end{table*}%

\begin{figure}[htbp]
    \centering
    \includegraphics[width=\linewidth]{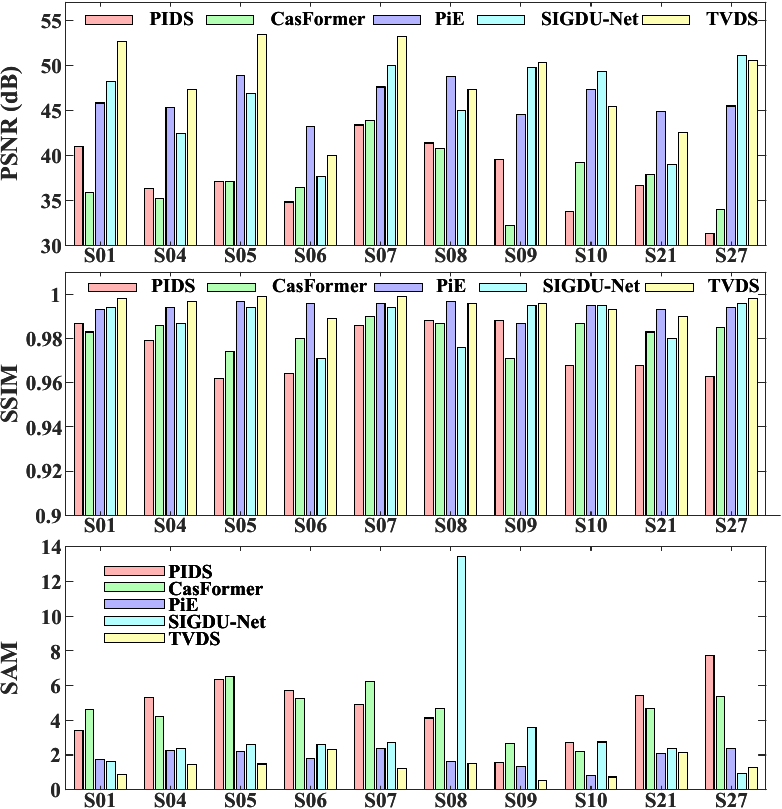}
    \caption{The PSNR, SSIM, and SAM comparison of different DC-CASSI methods on the CAVE-28 dataset}
    \label{Fig:Result-CAVE}
\end{figure}

Notably, TVDS maintains stable performance across diverse test scenes, as illustrated in \Cref{Fig:Result-CAVE}. 
TVDS maintains leading PSNR/SSIM and the smallest SAM across most scenes. {In contrast, SIGDU-Net exhibits notable SAM degradation in S08, as its spectral TV loss erroneously regularizes motion-induced spatial shifts, thereby distorting spectral signatures.}
% In contrast, SIGDU-Net shows notable SAM degradation in S08. 

% while the baseline Initialization method performs far worse, highlighting the efficacy of TVDS regularization.

\begin{figure}[htbp]
    \centering
    \includegraphics[width=\linewidth]{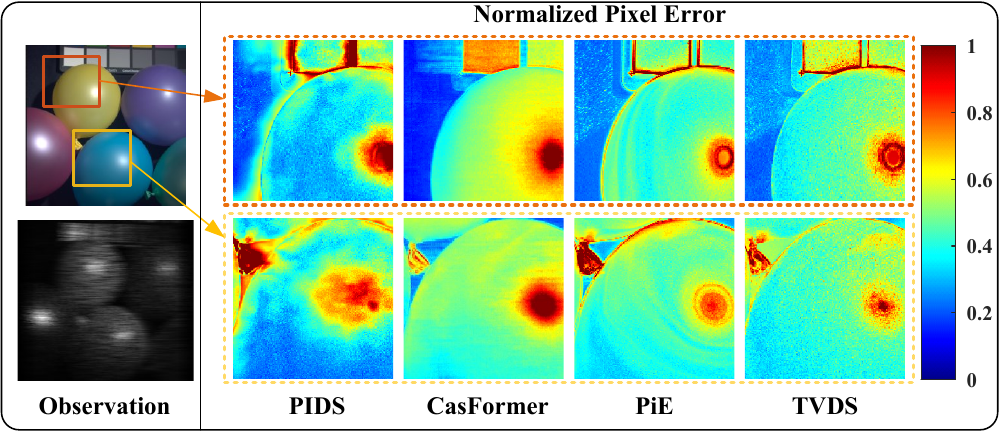}
    \caption{Visualization of normalized reconstruction errors for two distinct regions in CAVE-28 S01}
    \label{VIS:CAVE_01}
\end{figure}

Figure \ref{VIS:CAVE_01} further confirm this advantage: TVDS minimizes reconstruction errors in both edge and bright regions of S01, whereas PIDS/PiE exhibit edge error fluctuations (due to static/misaligned priors) and CasFormer has concentrated errors in bright regions. This validates that TVDS not only optimizes metrics but also delivers detail-rich reconstruction.
\begin{figure*}[!ht]
    \centering
    \includegraphics[width=\linewidth]{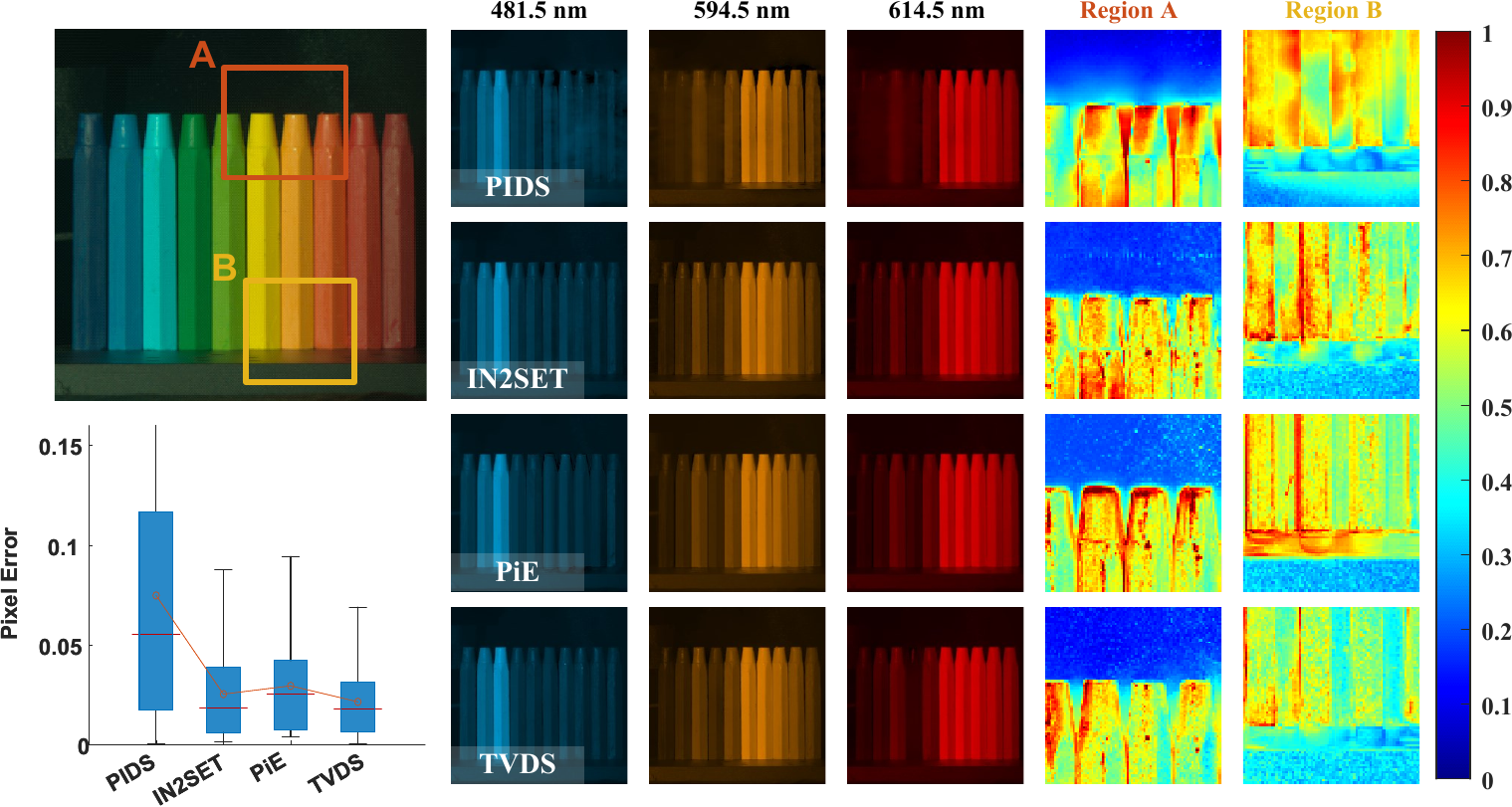}
    \caption{Visualization of TSA-Simu S09 scene reconstruction: Normalized region errors (right) and statistical pixel-wise error (bottom-left)}
    \label{VIS:KAIST_scene09}
\end{figure*}

\subsection{Comparison on TSA-Simu dataset}

As summarized in \Cref{Tab:TSA-simu}, TVDS leads comprehensively across the three core evaluation metrics, with notable advantages in average performance: it achieves an average PSNR of 46.94 dB, significantly outperforming key competitors like PiE and the attention-based deep unfolding network In2SET, and secures top performance in most test scenes—even its initialization result outperforms PIDS, a method also based on the model-based paradigm. In terms of SSIM, its average value of 0.994, outperforming In2SET and PiE overall and showing clear superiority over the baseline Initialization method. Regarding SAM, TVDS attains the lowest average of 2.996°, notably surpassing PiE and In2SET in effectively retaining critical spectral characteristics.

\Cref{VIS:KAIST_scene09} further confirms TVDS's advantages visually: in the normalized error maps of Regions A and B, TVDS exhibits the smallest error amplitude, whereas PIDS and In2SET show noticeable fluctuations in the inner crayon regions, and PiE has significantly higher errors at crayon tips (Region A) and reflection regions (Region B). The statistical pixel error distribution also shows TVDS has the narrowest range and lowest mean error (reflecting stable error suppression), while PIDS, In2SET, and PiE have wider distributions with more pixels deviating from the ground truth.

{
\subsection{Discussion on Noisy dataset}
\begin{table*}[htbp]
    
  \centering
  \caption{Noise robustness comparison on ARAD-Simu dataset. The average PSNR (dB), SSIM, and SAM (°) are reported.}
  \resizebox{\linewidth}{!}{
    \begin{tabular}{ccccccccccc}
        \toprule[1pt]
          &       & \multicolumn{3}{c}{Poor} & \multicolumn{3}{c}{Medium} & \multicolumn{3}{c}{Good} \\
          \cmidrule(lr){3-5}\cmidrule(lr){6-8}\cmidrule(lr){9-11}
    Methods & Metrics & 3:7   & 5:5   & 7:3   & 3:7   & 5:5   & 7:3   & 3:7   & 5:5   & 7:3 \\
    \midrule
    \multirow{3}[0]{*}{\makecell{PIDS \\ (2023, TPAMI)}} & PSNR  & 21.42  & 21.37  & 20.86  & 21.99  & 21.92  & 21.65  & 22.27  & 22.25  & 22.08  \\
          & SSIM  & 0.417  & 0.411  & 0.361  & 0.478  & 0.473  & 0.443  & 0.512  & 0.509  & 0.491  \\
          & SAM   & 43.253  & 42.862  & 45.180  & 39.894  & 39.277  & 40.514  & 37.703  & 37.323  & 37.977  \\
          \midrule
    \multirow{3}[0]{*}{\makecell{PiE \\ (2024, TCSVT)}} & PSNR  & 29.99  & 32.53  & 33.40  & 35.69  & 37.48  & 37.69  & 39.52  & 41.61  & 41.68  \\
          & SSIM  & 0.711  & 0.815  & 0.833  & 0.893  & 0.920  & 0.930  & 0.955  & 0.971  & 0.969  \\
          & SAM   & 20.187  & 13.400  & 11.531  & 9.706  & 7.331  & 6.958  & 6.336  & 4.993  & 4.777  \\
          \midrule
    \multirow{3}[0]{*}{\makecell{TVDS \\ (Ours)}} & PSNR  & \textbf{40.67 } & \textbf{40.23 } & \textbf{38.68 } & \textbf{43.15 } & \textbf{42.57 } & \textbf{41.13 } & \textbf{44.91 } & \textbf{44.39 } & \textbf{43.24 } \\
          & SSIM  & \textbf{0.960 } & \textbf{0.954 } & \textbf{0.933 } & \textbf{0.977 } & \textbf{0.973 } & \textbf{0.962 } & \textbf{0.985 } & \textbf{0.982 } & \textbf{0.977 } \\
          & SAM   & \textbf{6.450 } & \textbf{6.302 } & \textbf{7.403 } & \textbf{4.627 } & \textbf{4.828 } & \textbf{5.613 } & \textbf{3.824 } & \textbf{3.991 } & \textbf{4.487 } \\
          \bottomrule[1pt]
    \end{tabular}}
  \label{tab:noise_test}%
\end{table*}%
To evaluate noise robustness, experiments were conducted on the ARAD-Simu dataset spanning three imaging conditions (Poor/Medium/Good) and three light split ratios (3:7/5:5/7:3).

As shown in \Cref{tab:noise_test}, TVDS consistently outperforms competing methods across all configurations. While PIDS exhibits noise vulnerability and PiE shows limited adaptation to extreme noise despite self-supervised regularization, TVDS maintains superior performance through its adaptive RGB reference generator that filters noise while extracting spatial priors, and TVDS regularization that aligns spectral structural directions with RGB references to suppress noise propagation. The framework demonstrates stable performance advantages even under varying light split ratios and challenging imaging conditions.
}

{
\subsection{Ablation Study}
Ablation experiments are designed to verify two core advantages of the TVDS framework: the efficiency of adjoint operator derived from proposed observation model and the necessity of each component in our reconstruction pipeline.

To validate that the adjoint operator \(\mathbf{\Phi}^\top\) accelerates reconstruction without performance loss, we replaced In2SET's \(\mathbf{\Phi}^\top\) with our proposed version, testing on the TSA-simu dataset. Results are summarized in Table \ref{tab:property2_efficiency}.
\begin{table}[htbp]
  \centering
  
  \caption{Ablation study of proposed SD-CASSI observation model.}
  \resizebox{0.95\linewidth}{!}{%
  \begin{tabular}{cccccc}
    \toprule
    Implementation       & PSNR (dB) & SSIM  & SAM (°) & Time (ms) & \makecell{Time \\ of \(\mathbf{\Phi}^\top\) (ms)} \\
    \midrule
    In2SET    & 43.73     & 0.988 & 6.348    & 232.74                    & 42.5                                        \\
    Ours & 43.73     & 0.988 & 6.348    & \textbf{226.86}                    & \textbf{36.6}                                        \\
    \bottomrule
  \end{tabular}
  \label{tab:property2_efficiency}
  }
\end{table}

As shown in Table \ref{tab:property2_efficiency}, the two implementations achieve identical PSNR, SSIM, and SAM, confirming no performance loss. In terms of efficiency, total running time is reduced by $\sim$6 ms, with \(\mathbf{\Phi}^\top\) computation time dropping by 5.9 ms. This aligns with theoretical expectations (theoretical reduction should be {\(42.5\,\mathrm{ms} \times s(C-1)/W \approx 8.86\,\mathrm{ms}\)} for shear step \(s=2\), spectral bands \(C=28\), spatial width \(W=256\) on TSA-simu dataset), validating the \cref{Property:Efficient_Adjoint} of proposed observation model.

An incremental ablation study on CAVE-28 dataset validates the contribution of each component (TV subgradient guidance, reference update), as shown in Table \ref{tab:addlabel}.
\begin{table}[htbp]
  \centering
  \caption{Ablation study of proposed reconstruction framework.}
  \resizebox{\linewidth}{!}{
    \begin{tabular}{ccccccc}
    \toprule
    Initialization & TV & \makecell{Subgradient \\ guidance} & \makecell{Reference \\update} & PSNR & SSIM & SAM \\
    \midrule
\checkmark  &       &       &       & 34.73  & 0.778 & 9.253 \\
    \checkmark  & \checkmark  &       &       &{37.57}  & {0.928} & {4.671} \\
    \checkmark  & \checkmark  & \checkmark  &        & 47.55  & 0.995 & 1.433 \\
    \checkmark  & \checkmark  & \checkmark  &  \checkmark & 48.29 & 0.996 & 1.363 \\

    \bottomrule
    \multicolumn{7}{p{\linewidth}}{\footnotesize \textbf{Note:} TV-only baseline with its optimally tuned parameters ($\eta=3 \times 10^{-4}$, $\rho=1$, and $\beta=1.4$).}
    \end{tabular}%
  }
  \label{tab:addlabel}%
\end{table}%

The baseline (physics-informed initialization only) achieved modest performance (34.73 dB PSNR, 0.778 SSIM, 9.253° SAM). {Incorporating TV regularization with optimally tuned parameters improved all metrics (37.57~dB PSNR, 0.928 SSIM, 4.671$^\circ$ SAM), yet its inherent over-smoothing tendency limits fine detail preservation.} The integration of subgradient guidance yielded a dramatic performance leap, boosting PSNR to 47.55 dB and SSIM to 0.995, demonstrating its efficacy in mitigating over-smoothing while preserving details. Finally, the complete framework with dynamic reference update achieved optimal results (48.29 dB PSNR, 1.363° SAM), confirming that refined TV subgradients further improve reconstruction accuracy.

% Adding TV regularization alone degraded PSNR to 31.17 dB but improved spectral fidelity and structural similarity. Integrating subgradient guidance yielded dramatic improvements (47.55 dB PSNR, 0.995 SSIM, 1.433° SAM) by capturing local structural changes and mitigating TV's over-smoothing effect. The complete framework with dynamic reference update achieved optimal performance, demonstrating how adaptive alignment of TV subgradients with spectral image characteristics further refines reconstruction accuracy.
}

\subsection{Discussion on Computational Efficiency}
The computational efficiency of TVDS is evaluated by comparing its running time with PiE~\cite{PiE} and PIDS~\cite{ADMM_PIDS} across three datasets of varying spatial-spectral resolutions, as summarized in Table~\ref{tab:Running time}.
\begin{table}[htbp]
  \centering
  \caption{Running time comparison in different datasets.}
  \resizebox{\linewidth}{!}{
    \begin{tabular}{cccc}
    \toprule[1pt]
    Dataset & TVDS  & PiE   & PIDS \\
    \midrule
    TSA-simu (\(256\times256\times28\)) & \textbf{8.27 s}  & 33.08 s  & 11.14 s  \\
    ARAD-simu (\(256\times256\times31\)) & \textbf{9.40 s}  & 37.12 s  & 11.30 s  \\
    CAVE-28 (\(512\times512\times28\)) & 48.34 s  & 189.09 s  & \textbf{44.70 s}  \\
    \bottomrule[1pt]
    \multicolumn{4}{l}{\footnotesize\textbf{Note:} \textbf{Bold values} denote the minimum running time.}
    \end{tabular}%
    }
  \label{tab:Running time}%
\end{table}

{
TVDS achieves nearly 4.0$\times$ speedup over PiE across all benchmarks. On CAVE-28, PIDS' slight edge (44.70 s vs. 48.34 s) stems from parameter tuning that cuts TV iterations by 16.2\% vs. TSA-simu.

Moreover, the running time does not scale linearly with data volume, largely due to fixed overheads and memory access latency, as detailed in Supplementary Section S4.
}

\section{Experiment on Real Dataset}
\label{sec7}

\begin{figure*}[!htbp]
  \centering
  \renewcommand{\thefigure}{12}
  \includegraphics[width=0.9\linewidth]{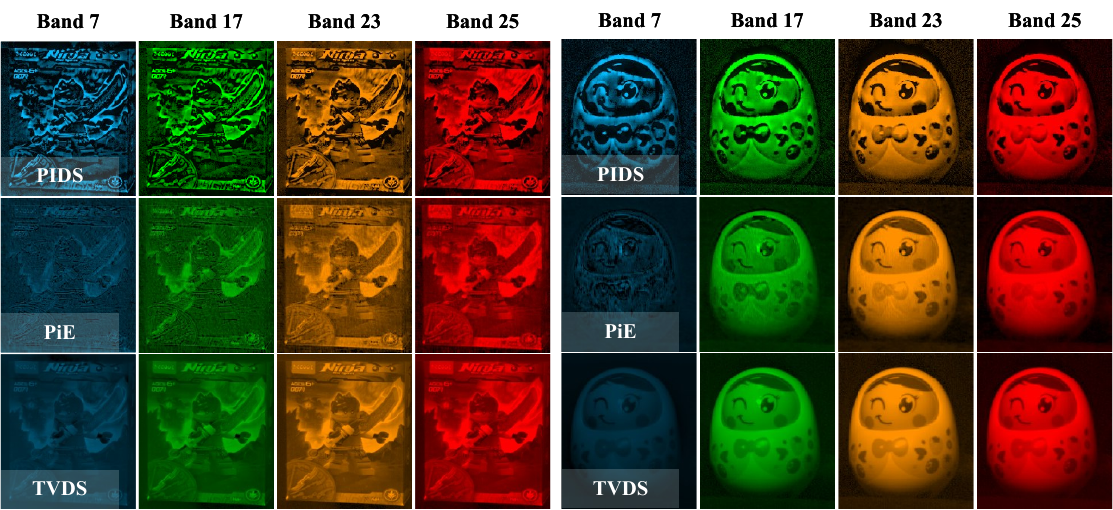}
  \caption{Comparative visualization of DC-CASSI reconstruction algorithms on the "Ninja" and "Doll" dataset}
  \label{Fig:real_test}
\end{figure*}
{
To verify the practical applicability of the proposed TVDS method, experiments were conducted on two real DC-CASSI scenes (“Ninja” and “Doll”) from \cite{rTVRA}. Each scene includes three key components: panchromatic (PAN) measurement, SD-CASSI measurement, and panchromatic sensor spectral response function {(SRF)}, as shown in \Cref{Fig:RealDataset}.
\begin{figure}[ht]
    \centering
    
    \renewcommand{\thefigure}{11}
    \includegraphics[width=0.95\linewidth]{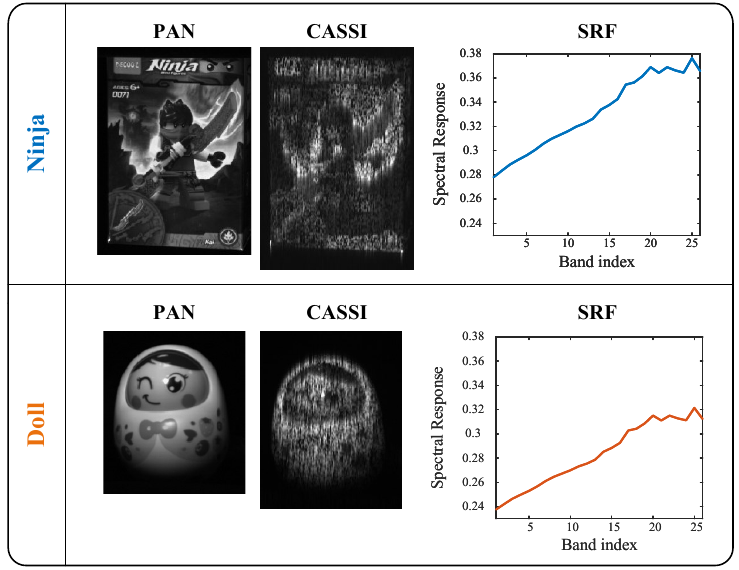}
    \caption{Visualization of the real dataset, including "Ninja" and "Doll" scenes}
    \label{Fig:RealDataset}
\end{figure}
{Due to the absence of ground-truth spectral images, we evaluate performance via PAN reprojection. To mitigate radiometric inconsistencies (e.g., exposure time, beam-splitting, normalization) that could invalidate the evaluation, we employ an adaptive matching strategy that calibrates CASSI observations via an estimated scaling factor $\kappa$ (see Supplementary Section S3 for details). For fair comparison, we benchmark TVDS against methods without explicit SRF constraints (PIDS \cite{ADMM_PIDS} and PiE \cite{PiE}), with results summarized in \Cref{Tab:real_test}.}
\begin{table}[!ht]
  \centering
  \caption{Quantitative Evaluation on Real Dataset}
  \resizebox{0.95\linewidth}{!}{
    \begin{tabular}{c l cccc}
        \toprule[1 pt]
        \textbf{Scene} & \textbf{Method} & \textbf{NIQE} & \textbf{BRISQUE} & \textbf{PSNR} & \textbf{SSIM} \\
        \midrule
        \multirow{4}{*}{\textbf{Ninja}}
        & PIDS  & 7.277 & 39.20 & 6.84 & 0.313 \\
        & PiE   & 7.351 & \textbf{31.64} & 19.18 & 0.505 \\
        & \textbf{TVDS} & \textbf{3.291} & 32.28 & \textbf{26.78} & \textbf{0.882} \\
        \cmidrule{2-6}
        & PAN input & 3.386 & 35.41 & - & - \\
        \midrule
        \multirow{4}{*}{\textbf{Doll}}
        & PIDS  & 8.710 & 27.81 & 9.44 & 0.420 \\
        & PiE   & 7.903 & 27.34 & 28.39 & 0.847  \\
        & \textbf{TVDS} & \textbf{7.212} & \textbf{22.29} & \textbf{34.50} & \textbf{0.973} \\
        \cmidrule{2-6}
        & PAN input & 7.373 & 20.80 & - & - \\
        \bottomrule[1pt]
        \multicolumn{6}{l}{\footnotesize\textbf{Note:} \textbf{Bold values} denote top performance.}
    \end{tabular}
  }
  \label{Tab:real_test}
\end{table}

As shown in \Cref{Tab:real_test}, TVDS's reconstruction quality is comparable to the original PAN input in terms of perceptual metrics (NIQE), while significantly outperforming existing methods in fidelity metrics (PSNR and SSIM). This highlights TVDS's ability to effectively fuse spatial and spectral information from real measurements. }

To complement quantitative results, \Cref{Fig:real_test} visualizes reconstructions of four spectral bands (Bands 7, 17, 23, 25) for both scenes—the left subfigure corresponds to “Ninja” and the right to “Doll”. PIDS reconstructions are heavily contaminated by noise (notably in the “Doll” scene); PiE retains spatial structure in Bands 23, 25 but loses detail in Band 7 and parts of Band 17 (e.g., the doll's hair). TVDS fully preserves the PAN input's spatial structure and has significantly lower noise in high-frequency bands than the other methods.

\section{Conclusion}
\label{sec8}
This study proposes a novel multi-source fusion framework for DC-CASSI reconstruction. By fusing compressive sensing-based CASSI measurements with spatial priors from auxiliary RGB/panchromatic image, it advances SOTA spectral reconstruction via innovations in architecture and algorithm.

Three interconnected innovations form the framework's core. First, the tensor-form Kronecker $\delta$-based end-to-end SD-CASSI observation model establishes a rigorous, efficient mathematical foundation for reconstruction. Second, the adaptive reference generation mechanism enables dynamic spatial prior extraction and ensures compatibility across diverse sensor setups. Third, the TVDS regularization term innovates reconstruction algorithm by encoding structural direction information from references into spectral reconstruction.

Comprehensive validations on simulated and real DC-CASSI datasets confirm the framework's superiority. In performance, it outperforms SOTA methods on key metrics (PSNR, SSIM, SAM) and exhibits robust noise resistance under challenging imaging conditions (high ISO, short exposure, varying light split ratios). In efficiency, its linear complexity scaling ensures applicability to high-resolution spectral images. Ablation studies validate each component's effectiveness and confirm the integrated framework fully leverages cross-sensor complementarity.

This work contributes to snapshot spectral imaging by establishing an interpretable, efficient fusion-based reconstruction framework. It enriches the theoretical and algorithmic toolkit for multi-sensor computational imaging and can be extended to other complementary image/video processing scenarios.

% \bibliographystyle{IEEEtran.bst}
% \bibliography{IEEEabrv, ref.bib}

\end{document}

\title{Supplementary Material for \\ ``TV Subgradient-Guided Multi-Source Fusion for Spectral Imaging in Dual-Camera CASSI Systems''}

\author{Weiqiang Zhao\orcidlink{0009-0009-2714-4622},
Tianzhu Liu\orcidlink{0000-0001-6903-9614},
Yuzhe Gui\orcidlink{0009-0007-2630-6736},
Wei Bian\orcidlink{0000-0003-4252-047X},
Yanfeng Gu\(^*\)\orcidlink{0000-0003-1625-7989}
}
\maketitle

{
    \hypersetup{linkcolor=black}
    \tableofcontents
}
\begin{minipage}{\textwidth}\ \\[30pt] \footnotesize \centering 
Copyright \copyright 2026 IEEE. Personal use of this material is permitted. \\
However, permission to use this material for any other purposes must be obtained from the IEEE by sending an email to pubs-permissions@ieee.org.
\end{minipage}
\newpage
\section{Spatial alignment}
\subsection{Methodology}
In practice, spatial alignment can be implemented with hardware calibration and algorithmic alignment. For hardware calibration,  we can follow the calibration process described in \cite{DC-CASSI_pan}. 

However, residual misalignment may still exist due to mechanical vibrations, temperature variations, or imperfect calibration. To address this, we introduce a learnable module. Considering the spatial transformation operation for alignment possesses translation invariance, which aligns with the prior of 2D convolution. Moreover, since the alignment task does not involve deep semantic understanding, the deep network is unnecessary, and the pixel offset is small, a large receptive field is unnecessary. Therefore, we use a simple inception-like network to effectively achieve alignment.

In order to train the network in an unsupervised way, we use the CASSI observation prior to guide the spatial alignment, which makes the reprojected CASSI from RGB spatially aligned with the real measurements.

\begin{figure}[htbp]
    \centering
    \includegraphics[width=0.8\textwidth]{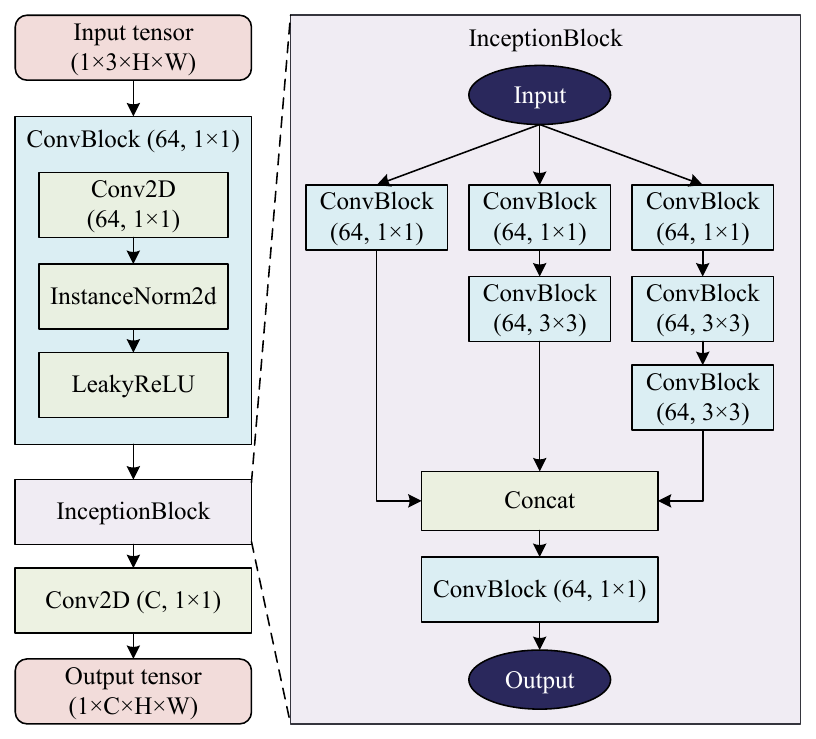}
    \caption{Architecture of the spatial alignment network.}
    \label{fig:net}
\end{figure}

Therefore, the network architecture is illustrated in Fig.~\ref{fig:net}.

The network parameters are optimized by minimizing the measurement consistency loss:
\begin{equation}
    \min_{\boldsymbol{\theta}} \quad \|\boldsymbol{y} - \mathbf{\Phi}\,\mathrm{vec}(f(\mathcal{Y}_{\text{rgb}}; \boldsymbol{\theta}))\|_1,
    \label{eq:reg_loss}
\end{equation}
where $\boldsymbol{y}$ denotes the CASSI compressed measurements, $\mathbf{\Phi}$ represents the sensing matrix, and $f(\cdot; \boldsymbol{\theta})$ is the alignment network with parameters $\boldsymbol{\theta}$. This loss ensures that the output is spatially consistent with CASSI measurements.

\subsection{Simulated Experiment}
To validate the effectiveness of the proposed alignment method, we compare the reconstruction performance with and without the alignment network under various pixel shift conditions based on the TSA-simu dataset. We set the negative slope of LeakyReLU 0.2, and train the network with Adam optimizer with learning rate $10^{-3}$ for 1000 epochs. The results are summarized in Table~\ref{Tab:Alignment}.

\begin{table}[htbp]
  \centering
  \caption{Quantitative comparison of reconstruction performance with and without the alignment network under different pixel shift conditions.}
    \begin{tabular}{ccccccc}
    \toprule
          & \multicolumn{3}{c}{With Network} & \multicolumn{3}{c}{Without Network} \\
    Pixel Shift & PSNR  & SSIM  & SAM   & PSNR  & SSIM  & SAM \\
    \midrule
    0     & -     & -     & -     & 46.94  & 0.994  & 2.996  \\
    $0.25\sqrt{2}$ & 46.24  & 0.994  & 3.039  & 41.76  & 0.988  & 3.728  \\
    0.5   & 45.28  & 0.993  & 3.119  & 39.69  & 0.982  & 4.188  \\
    $0.5\sqrt{2}$ & 44.46  & 0.992  & 3.281  & 37.42  & 0.972  & 5.029  \\
    \bottomrule
    \end{tabular}
  \label{Tab:Alignment}
\end{table}

As shown in Table~\ref{Tab:Alignment}, the alignment network consistently improves reconstruction quality across all misalignment levels. Without the network, performance degrades significantly as pixel shift increases (PSNR drops from 46.94 to 37.42 dB). In contrast, with the alignment network, the degradation is substantially mitigated (PSNR only drops from 46.24 to 44.46 dB). This demonstrates the effectiveness of our algorithmic refinement in compensating for residual spatial misalignment.

\section{Parameter Tuning for TV-only Baseline}
To ensure that the performance improvements attributed to subgradient guidance are not confounded by inadequate baseline tuning, we performed hyperparameter searches over three critical parameters: $\eta$ (i.e., \(\mu/\rho\)), $\rho$, and $\beta$. 

\begin{figure}[htbp]
    \centering
    \includegraphics[width=\textwidth]{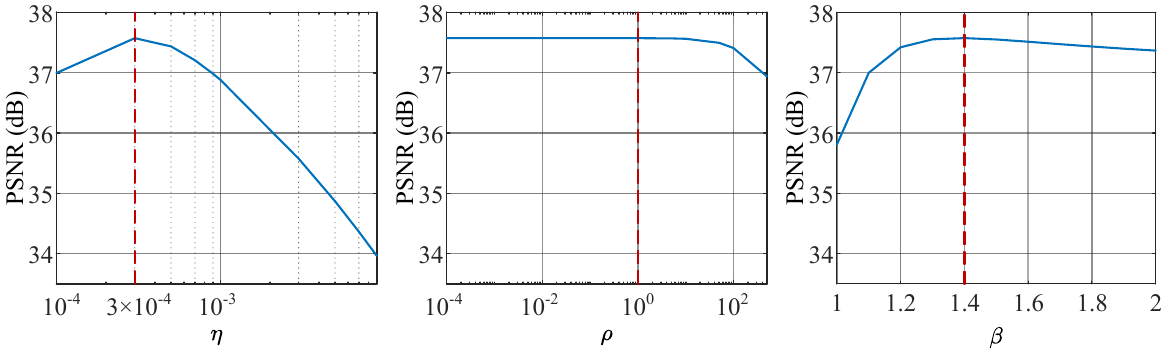}
    \caption{Parameter sweep results to verify the optimality of the TV-only baseline configuration. (a) Varying $\eta$; (b) Varying $\rho$; (c) Varying $\beta$. The red vertical dashed line marks the optimal configuration with the highest PSNR.}
    \label{fig:tv_sweep}
\end{figure}

All experiments were performed on the CAVE-28 dataset, with optimal hyperparameters selected based on the PSNR metric. We employed a sequential single-parameter sweeping strategy (fixing the other two hyperparameters to their optimal values determined in prior steps) to identify the optimal configuration. The final optimal hyperparameters were identified as \(\eta=3 \times 10^{-4}\), \(\rho=1\), and \(\beta=1.4\). We further verified this configuration by sweeping the parameters in its neighborhood, as illustrated in Fig.~\ref{fig:tv_sweep}. 

\section{Radiometric Matching for Real Dataset Evaluation}
\subsection{Methodology}
Radiometric matching is a prerequisite for valid re-projection-based evaluation. As the spectral estimate obtained in Main Document Sec.~IV-B is energy-consistent solely with the CASSI branch, we employ a three-step radiometric matching procedure:
\begin{enumerate}[label=(\arabic*)]
    \item \textbf{Initial Projection:} Replace the RGB input of the method in Sec.~IV-B with the PAN image to generate a spectral estimate, then project it to the PAN domain via the sensor SRF to obtain \(\hat{\mathbf{Y}}_{\text{pan}}\).
    \item \textbf{Coefficient Estimation:} Compute the alignment coefficient $\kappa$ by minimizing the MSE between \(\hat{\mathbf{Y}}_{\text{pan}}\) and the measured PAN image \(\mathbf{Y}_{\text{pan}}\):
    \begin{equation}
        \kappa = \frac{\langle\hat{\mathbf{Y}}_{\text{pan}}, \mathbf{Y}_{\text{pan}}\rangle}{\|\hat{\mathbf{Y}}_{\text{pan}}\|_F^2}. \label{eq:kappa_est}
    \end{equation}
    \item \textbf{Observation Calibration:} Last, we get calibrated CASSI measurements $\mathbf{Y}^{\text{cal}}_{\text{cassi}} = \kappa \mathbf{Y}^{\text{orig}}_{\text{cassi}}$. This ensures that the CASSI branch is energy-matched to the PAN branch. Consequently, the reconstructed spectral image satisfies the requirements for reprojection-based evaluation.
\end{enumerate}

\subsection{Simulated Experiment}
To evaluate estimation accuracy, we conducted Monte Carlo simulations configured with real-system parameters (coded aperture and SRF) from the ``Ninja'' and ``Doll'' scenes \cite{rTVRA}. The simulation protocol was structured as follows:
\begin{itemize}
    \item \textbf{Noise Model:} Additive white Gaussian noise (distributed as $\mathcal{N}(0, \sigma^2 \mathbf{I})$) was added to the measurements, with noise standard deviations $\sigma \in \{1.0, 0.3, 0.1, 0.03, 0.01\}$.
    \item \textbf{Data Generation:} For each of 1,000 trials per $\sigma$, the ground-truth energy ratio $\kappa_{\text{gt}}$ was sampled uniformly from $[0.1, 10]$. Hyperspectral cubes with i.i.d. uniform entries $\in [0,1]$ were generated to represent a worst-case scenario lacking spatial-spectral correlation.
    \item \textbf{Evaluation Metric:} Accuracy was quantified via the Normalized Mean Squared Error (NMSE) in dB:
    \begin{equation}
        \text{NMSE}_{\kappa} = 10 \log_{10} \left( \mathbb{E} \left[ \frac{(\hat{\kappa} - \kappa_{\text{gt}})^2}{\kappa_{\text{gt}}^2} \right] \right). 
    \end{equation}
\end{itemize}

Table~\ref{tab:kappa_estimation_nmse} summarizes the NMSE performance of the estimated radiometric alignment coefficient $\hat{\kappa}$ across different noise levels and scenes.

\begin{table}[htbp]
  \centering
  \caption{NMSE (dB) of the estimated radiometric alignment coefficient $\hat{\kappa}$ under varying noise levels $\sigma$.}
  \label{tab:kappa_estimation_nmse}
  \begin{tabular}{lccccc}
    \toprule 
    & \multicolumn{5}{c}{\textbf{Noise Std. $\sigma$}} \\
    \cmidrule(lr){2-6}
    \textbf{Scene} & \textbf{1.0} & \textbf{0.3} & \textbf{0.1} & \textbf{0.03} & \textbf{0.01} \\
    \midrule
    Ninja & -27.36 & -48.17 & -61.81 & -67.23 & -68.11 \\
    Doll & -29.93 & -48.93 & -60.96 & -66.16 & -66.70 \\
    \bottomrule
    \multicolumn{6}{l}{{\footnotesize Lower NMSE indicates higher estimation accuracy.}}
  \end{tabular}
\end{table}
The results exhibit three key characteristics. First, estimation accuracy improves monotonically with SNR; reducing $\sigma$ from 1.0 to 0.1 yields an $\approx 31$ dB NMSE improvement for both scenes, confirming that the estimator in Eq.~(\ref{eq:kappa_est}) effectively leverages high-quality measurements. Second, performance saturates at $\approx -66$ dB for $\sigma \leq 0.03$. This error floor is attributed to the rank-1 subspace approximation in the closed-form initialization (Sec. IV-B), which introduces noise-independent approximation error. Finally, the consistent trends across ``Ninja'' and ``Doll'' (variations within 3 dB) demonstrate the method's robustness to varying system configurations.

\section{Discussion of Running Time}
\label{sec:runtime_analysis}

\subsection{Theoretical Analysis}
\label{subsec:theoretical_model}

The total running time $T(N)$ for processing a spectral image of volume $N$ (measured in millions of elements) is decomposed into three components:
\begin{equation}
    T(N) = T_{\text{fix}} + T_{\text{comp}}(N) + T_{\text{mem}}(N),
    \label{eq:total_time}
\end{equation}
where $T_{\text{fix}}$ denotes fixed overhead (e.g., initialization), $T_{\text{comp}}(N)$ represents computation time, and $T_{\text{mem}}(N)$ accounts for hierarchical memory access latency.

The computation time scales linearly with $N$, as established in Section~V-D:
\begin{equation}
    T_{\text{comp}}(N) = c_{\text{comp}} N,
    \label{eq:comp_time}
\end{equation}
where $c_{\text{comp}}$ is the computational cost per million elements. Memory access time is modeled as a weighted combination of cache-hit and cache-miss latencies:
\begin{equation}
    T_{\text{mem}}(N) = N \big[ c_{\text{cache}}  H(N) + c_{\text{mem}} (1 - H(N)) \big],
    \label{eq:mem_time}
\end{equation}
where $c_{\text{cache}}$ and $c_{\text{mem}}$ denote the access costs per million elements for cache and global memory, respectively, and $H(N) \in [0,1]$ is the cache hit rate.

Following the classical stack distance theory~\cite{Stack_Distance}, the cache hit rate $H(N;C)$ under the LRU assumption corresponds to the cumulative distribution function (CDF) of the stack distance evaluated at the cache capacity $C$. In our proposed fixed-point iterations, the memory access pattern is characterized by frequent alternating accesses between variables $\mathcal{Z}$ and $\mathcal{P}_{\mathcal{Z}}$. This results in a stack distance distribution that is tightly concentrated. To capture this characteristic, we model the stack distance using a log-logistic distribution, yielding the following hit rate formulation:
\begin{equation*}
    H(N) = \frac{1}{1 + (\alpha N / C)^{\beta}},
\end{equation*}
where $C$ is the physical cache capacity, $\alpha$ scales the algorithm's working set relative to $N$, and $\beta > 0$ characterizes access locality. By defining the effective cache capacity $C_{\text{eff}} \triangleq C / \alpha$, we obtain the simplified form:
\begin{equation}
    H(N) = \frac{1}{1 + (N / C_{\text{eff}})^{\beta}}.
    \label{eq:hit_rate}
\end{equation}
This formulation effectively captures the gradual degradation of cache efficiency as the working set exceeds hardware limits.

\subsection{Experimental Validation}
\label{subsec:experimental_validation}

We fitted the proposed model to empirical running time measurements using nonlinear least-squares optimization. Table~\ref{tab:parameters} summarizes the fitted parameters.

\begin{table}[htbp]
\centering
\caption{Fitted model parameters and goodness of fit.}
\label{tab:parameters}
\begin{tabular}{lcr}
\toprule
\textbf{Parameter} & \textbf{Symbol} & \textbf{Value} \\
\midrule
Cache access cost & $c_{\text{cache}}$ & $2.98 \times 10^{-2}$ (s/M-elements) \\
Global memory access cost & $c_{\text{mem}}$ & $6.33$ (s/M-elements) \\
Effective cache capacity & $C_{\text{eff}}$ & $4.07$ (M-elements) \\
Transition sharpness & $\beta$ & $3.07$ \\
Computation cost & $c_{\text{comp}}$ & $6.85 \times 10^{-7}$ (s/M-elements) \\
Fixed overhead & $T_{\text{fix}}$ & $6.65$ (s) \\
\midrule
\multicolumn{2}{r}{\textbf{Goodness of fit ($R^2$):}} & $0.9999$ \\
\bottomrule
\end{tabular}
\end{table}

\begin{figure}[t!]
    \centering
    \includegraphics[width=0.85\textwidth]{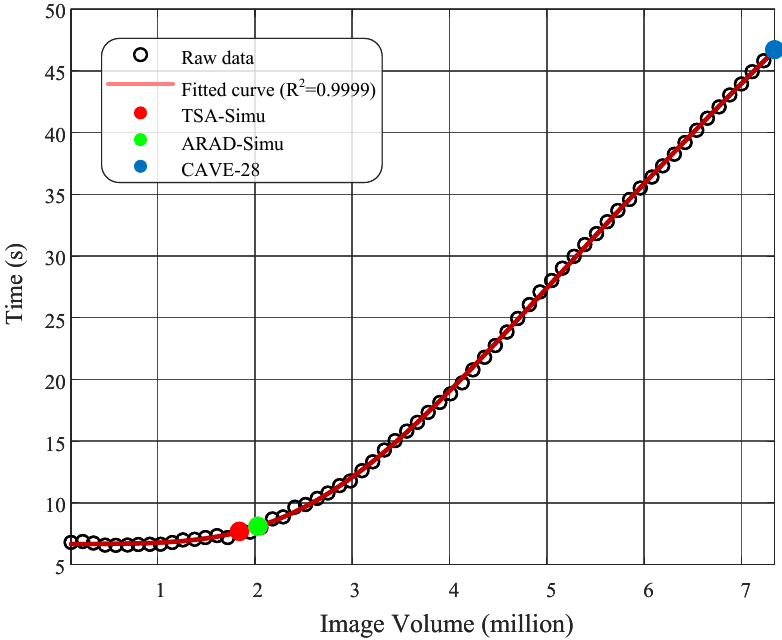}
    \caption{Running time versus data volume: raw measurements (markers) versus fitted model (solid curve).}
    \label{fig:fit_curve}
\end{figure}

The near-perfect coefficient of determination ($R^2 = 0.9999$) confirms that the model accurately captures the relationship between observed running time and data volume, as visually validated in Fig.~\ref{fig:fit_curve}.

The fitted parameters reveal several key characteristics of the system performance:

\begin{enumerate}[label=(\arabic*)]
    \item \textbf{Memory-Bound Nature:} The negligible computation cost ($c_{\text{comp}} \ll c_{\text{mem}}$) confirms that the algorithm is memory-bound. Execution time is dominated by data movement rather than arithmetic operations.
    \item \textbf{Locality Sensitivity:} The $\sim$212$\times$ latency gap between global memory and cache access ($c_{\text{mem}} / c_{\text{cache}}$) underscores the critical importance of data locality for performance optimization.
    \item \textbf{Hardware Alignment:} The effective cache capacity $C_{\text{eff}} \approx 4.07$\,M elements aligns closely with the RTX 5070 Ti's 48\,MB L2 cache. This assumes a $3\times$ working set expansion (accounting for $\mathcal{Z}$ and $\mathcal{P}_{\mathcal{Z}}$) with float32 precision ($4.07 \times 10^6 \times 4\,\text{B} \times 3 \approx 48.8\,\text{MB}$).
    \item \textbf{Overhead Dominance:} The fixed overhead ($T_{\text{fix}} \approx 6.65$\,s) implies that constant costs dominate total running time for very small problem sizes.
\end{enumerate}

These observations motivate dataset-specific optimization strategies. For small-scale inputs ($N < C_{\text{eff}}$, e.g., TSA-Simu and ARAD-Simu), reducing fixed overhead yields the greatest benefits. Conversely, for large-scale inputs ($N > C_{\text{eff}}$, e.g., CAVE-28), maximizing memory throughput via coalesced access patterns becomes the primary optimization target.

In summary, the derived model not only fits empirical data with exceptional fidelity but also provides hardware-aware guidance for optimizing GPU-based spectral reconstruction algorithms across varying problem scales.

% \bibliographystyle{IEEEtran}
% \bibliography{ref}